\documentclass[11pt]{article}

\usepackage[margin=1.1in]{geometry}
\usepackage{amsmath,amssymb,amsthm}
\usepackage{mathtools}
\usepackage{booktabs}
\usepackage{enumitem}
\usepackage{microtype}
\usepackage{xcolor}
\usepackage[colorlinks=true,linkcolor=blue!55!black,citecolor=blue!55!black,urlcolor=blue!55!black]{hyperref}

\theoremstyle{plain}
\newtheorem{theorem}{Theorem}
\newtheorem{lemma}[theorem]{Lemma}
\newtheorem{proposition}[theorem]{Proposition}
\newtheorem{corollary}[theorem]{Corollary}
\newtheorem{definition}[theorem]{Definition}
\newtheorem*{theorem*}{Theorem}
\theoremstyle{definition}
\newtheorem{remark}[theorem]{Remark}
\newcounter{algorithm}

\newcommand{\X}{\mathcal{X}}
\newcommand{\R}{\mathbb{R}}
\newcommand{\Ex}{\mathbb{E}}
\newcommand{\TV}{\mathrm{TV}}
\newcommand{\Nflip}{N_{\mathrm{flip}}}
\newcommand{\Sone}{\hyperref[cond:S1]{(S1)}}
\newcommand{\Stwo}{\hyperref[cond:S2]{(S2)}}
\DeclareRobustCommand{\OS}{\hyperref[oracle:S]{(S)}}
\DeclareRobustCommand{\OC}{\hyperref[oracle:C]{(C)}}
\DeclareRobustCommand{\OP}{\hyperref[oracle:P]{(P)}}

\title{Statistically Optimal and Near Linear Time Hypothesis Selection}
\author{Satyen Kale\\Apple \and Amirreza Shaeiri\thanks{The work was done while this author was an intern at Apple.}\\Purdue University
\and Kunal Talwar\\Apple}
\date{}

\begin{document}
\maketitle\begin{abstract}
In hypothesis selection we are given $n$ distributions $q_1,\dots,q_n$ on a countable domain $\X$ and samples from an unknown
distribution $p$. The goal is a distribution $q$ with $\TV(p,q)\le\alpha\min_i\TV(p,q_i)+\epsilon$. The factor $\alpha=2$ is
optimal for algorithms using a number of samples from $p$ that is independent of $|\X|$, and in general only improper learners achieve it. We give a learner that achieves $\alpha=2$ with
$O(\log(n/\delta)/\epsilon^2)$ samples of $p$, with probability at least $1-\delta$. This matches the sample complexity of
Yatracos's factor-$3$ estimator, and it removes the logarithmic factors in the factor-$2$ bound of Bousquet et al.\ (FOCS 2021). The
learner accesses the $q_i$ only through sampling and pointwise comparisons. We measure its running time by its oracle
complexity, the number of oracle calls. In expectation it makes $O((n\epsilon^{-2}+\epsilon^{-4}\log\frac1\epsilon)\log\frac n\delta)$ oracle calls, independent
of $\X$. It runs
multiplicative weights over the constraints $\TV(q,q_i)\le\TV(p,q_i)$ and uses at most one sample of $p$ per round.
\end{abstract}

\section{Introduction}\label{sec:intro}

Hypothesis selection is a basic problem in distribution learning. We are given a finite class
$\mathcal Q=\{q_1,\dots,q_n\}$ of distributions on a domain $\X$, and samples from an unknown distribution $p$ on $\X$.
We take $n\ge1$, $\epsilon\in(0,1]$ and $\delta\in(0,1)$.
We want a distribution $q$ that is nearly as close to $p$ as the best member of the class:
\begin{equation}\label{eq:hs}
  \TV(p,q)\;\le\;\alpha\cdot\mathrm{OPT}+\epsilon,\qquad \mathrm{OPT}=\min_{i\in[n]}\TV(p,q_i),
\end{equation}
with probability at least $1-\delta$. The two figures of merit are the approximation factor $\alpha$ and the number of
samples from $p$. Neither may depend on $\X$, which can be infinite. So in general we cannot estimate the distances $\TV(p,q_i)$
themselves.

\paragraph{Prior work.} The classical approach goes back to Yatracos~\cite{Yat85} and is developed in the book of
Devroye and Lugosi~\cite{DL01}. It looks at $p$ only through the Scheff\'e sets $A_{ij}=\{x:q_i(x)>q_j(x)\}$. There are
at most $n^2$ of them, so $O(\log(n/\delta)/\epsilon^2)$ samples estimate $p$ on all of them, and the minimum distance
estimate then achieves $\alpha=3$. This estimate is \emph{proper}: it outputs a member of $\mathcal Q$. Mahalanabis and
\v{S}tefankovi\v{c}~\cite{MS08} showed that $3$ is the best factor for a broad family of proper algorithms, and Bousquet,
Kane and Moran~\cite{BKM19} showed that no proper algorithm does better in general (the revised proof corrects a gap pointed out in~\cite{AACS26}). \emph{Improper} algorithms do better, and the
same authors achieved $\alpha=2$ for every finite class~\cite{BKM19}. This factor is optimal, since no
algorithm with $\alpha<2$ has a sample complexity independent of $\X$, by a construction of Chan, Diakonikolas, Servedio
and Sun~\cite{CDSS14} (see~\cite[\S1.7]{BKM19}). The sample complexity of the factor-$2$ algorithms of~\cite{BKM19} is
either $\tilde O(\sqrt n/\epsilon^{5/2})$ or depends on $\log|\X|$. Bousquet, Braverman, Efremenko, Kol and
Moran~\cite{BBEKM21} reduced it to $\tilde O(\log n/\epsilon^2)$ for constant $\delta$.

Both factor-$2$ results start from one observation~\cite[Lemma~3]{BKM19}. Suppose that a distribution $\bar u$ is at least
as close to every $q_i$ as $p$ is, up to $\epsilon$:
\begin{equation}\label{eq:goal}
  \TV(\bar u,q_i)\;\le\;\TV(p,q_i)+\epsilon\qquad(i=1,\dots,n).
\end{equation}
Let $q_{i^\ast}$ be closest to $p$. The triangle inequality gives $\TV(p,\bar u)\le\TV(p,q_{i^\ast})+\TV(q_{i^\ast},\bar u)\le2\,\mathrm{OPT}+\epsilon$.
The distribution $p$ itself satisfies~\eqref{eq:goal}, but we do not know it.

\paragraph{Concurrent and independent work.} During the preparation of this manuscript, we became aware of the concurrent work of
Aamand, Aliakbarpour, Chen, Silwal, Stevens and Zou~\cite{AACSSZ26}, which studies the same problem.
All results presented in this manuscript were obtained independently and established before the concurrent preprint became publicly available.
We refer the reader to \hyperref[par:concurrent]{Subsection~\ref*{sec:results}} for a comprehensive discussion.

\paragraph{Computation.} The factor-$2$ algorithms of~\cite{BKM19,BBEKM21} are information-theoretic. The running time of
the algorithm of~\cite{BBEKM21} grows polynomially with both $n$ and $|\X|$, as its authors point out~\cite[\S1.3]{BBEKM21}. It
needs membership in the Scheff\'e sets and further operations on $\X$ that can be implemented efficiently only for
restricted classes. For factor $3$ and larger, efficient algorithms are known. Mahalanabis and
\v{S}tefankovi\v{c}~\cite{MS08} match the guarantee of the minimum distance estimate with a number of computations linear
in $n$, after preprocessing the class. Daskalakis and Kamath~\cite{DK14} give a tournament that uses $O(\log n/\epsilon^2)$
samples and $O(n\log n/\epsilon^2)$ time when some $q_i$ is $\epsilon$-close to $p$. Aliakbarpour, Bun and Smith~\cite{ABS24} gave the
first algorithm with factor $3$, $O(\log(n/\delta)/\epsilon^2)$ samples and time nearly linear in $n$, namely
$\tilde O(n/(\delta^3\epsilon^3))$ as stated in~\cite{AACS26}. Aamand, Aliakbarpour, Chen and Silwal~\cite{AACS26} improved the time to
$\tilde O(n/(\delta\epsilon^2))$, and to $O(n\log^3(n/\delta)/\epsilon^2)$ when the value of $\mathrm{OPT}$ is known in advance.
Zou~\cite{Zou26} gives a factor-$3$ algorithm with time $O(n\log(n/\delta)/\epsilon^2)$. These
algorithms are proper, so factor $3$ is the best they can achieve. Their model charges unit time for a sample of $p$ or
of any $q_i$ and for a query of a value $q_i(x)$. For mixtures of the $q_i$, no fixed expected approximation factor below $3$
works for all $n$ with sample complexity independent of $\X$~\cite[Corollary~3.3]{AACS26}. Aamand et al.\ ask whether some improper algorithm uses $O(\log(n/\delta)/\epsilon^2)$ samples, achieves a
factor below $3$, and runs in time that does not grow with $|\X|$~\cite[Question~1]{AACS26}.

\paragraph{The oracle model.} We have three oracles and nothing else:
\begin{itemize}
\item[(S)]\phantomsection\label{oracle:S} \textbf{sampling:} given $i$, return a fresh draw $Y\sim q_i$;
\item[(C)]\phantomsection\label{oracle:C} \textbf{comparison:} given $x$, $i$ and $j$, return whether $q_i(x)>q_j(x)$;
\item[(P)]\phantomsection\label{oracle:P} \textbf{$p$-sampling:} return a fresh draw $X\sim p$.
\end{itemize}
A comparison is a membership query for a Scheff\'e set. No oracle ever returns a number $q_i(x)$ or $p(x)$. Our
algorithms manipulate only sampled points, comparison bits, and weights that they maintain themselves. The output is a
succinct description of a distribution on $\X$, a list of numbers whose length does not depend on $\X$.
We take $\X$ countable to keep measure
theory out of the way. We count storage in real numbers. In asymptotic bounds we write $\log x$ for $\max\{1,\ln x\}$, and $\tilde O(h)$ hides a factor $\log^kh$ for a constant $k$.

\subsection{Our results}\label{sec:results}

\begin{theorem*}[Informal]
There is an algorithm that makes $O(\log(n/\delta)/\epsilon^2)$ calls to \OP{} and returns a succinct description of a
distribution $\bar u$ on $\X$, given by $O(n)$ numbers for each of $T=O(\log(n/\delta)/\epsilon^2)$ rounds. With probability at
least $1-\delta$, $\TV(\bar u,q_i)\le\TV(p,q_i)+2\epsilon$ for every $i$, and hence $\TV(p,\bar u)\le2\,\mathrm{OPT}+2\epsilon$. In
expectation the algorithm makes
\[
  O\Bigl(\min\{n\epsilon^{-2},\,\epsilon^{-3}\}\log\tfrac n\delta\Bigr)\ \text{calls to \OS{} and}\ 
  O\Bigl(\min\{n^2\epsilon^{-2},\,\epsilon^{-4}\log\tfrac1\epsilon+n\epsilon^{-2}\}\log\tfrac n\delta\Bigr)\ \text{calls to \OC{}}.
\]
For any $\beta\in(0,1)$ the same description gives a sampler that uses only \OS{} and \OC{}.
On the same event, its output law $\mu$ satisfies $\TV(\mu,q_i)\le\TV(p,q_i)+2\epsilon+\beta$ for every $i$.
\end{theorem*}

The algorithm is one framework, multiplicative weights over the $n+1$ constraints, that works with four randomized
schemes for estimating the gains fed into multiplicative weights: uniform proposals, DKW-sparsified weights, a window kernel on ranks, and a binomial
kernel with early rejection. All four use at most $T=O(\log(n/\delta)/\epsilon^2)$ samples of $p$. They differ in the number
of samples they draw from the $q_i$ and in the number of comparisons, as Table~\ref{tab:cost} shows. The theorem uses
the uniform scheme for $n\le1/\epsilon$. For larger $n$ it uses whichever of the window and binomial schemes has the smaller comparison bound.

The description lists, for each round, its weights and, for the DKW scheme, its empirical weights. Corollary~\ref{cor:master}
and Proposition~\ref{prop:normsafe} prove the bound on $\bar u$. The sampler is that of Lemma~\ref{lem:output},
and its extra $\beta$ comes only from its cap on the attempts. Taking $\epsilon$ and $\beta$ each equal to one third of the
requested additive error gives~\eqref{eq:hs} for the sampler. Section~\ref{sec:cost} gives the costs. So the algorithm
achieves the optimal factor $2$ with $O((\log n+\log(1/\delta))/\epsilon^2)$ samples of $p$. This is the sample complexity of the factor-$3$
minimum distance estimate~\cite{Yat85,DL01}. For $n\ge2$ and small $\epsilon,\delta$ the bound is necessary even when $\mathrm{OPT}=0$.
Take a binary code of $n$ words $\theta\in\{-1,1\}^k$ with relative distance at least $1/4$ and $k=O(\log n)$. Let $\mathcal Q$ consist of the laws $p_\theta(j,\pm)=(1\pm12\epsilon\theta_j)/(2k)$ on $\X=[k]\times\{+,-\}$, and let $p\in\mathcal Q$.
Distinct laws are at least $3\epsilon$ apart in TV and have KL divergence $O(\epsilon^2)$, so an $\epsilon$-accurate estimate identifies the word and Fano's inequality gives the $\log n$ term.
Testing Bernoulli laws with means $\frac12\pm2\epsilon$ gives the $\log(1/\delta)$ term.
The costs in \OS{} and \OC{} do not depend on $\X$. Each sample of $p$
is used once, so the analysis needs no tools from adaptive data analysis.

\paragraph{Running time.} We measure running time by oracle complexity, the number of calls to \OS{}, \OC{} and
\OP{}. In expectation the binomial scheme makes $O((n\epsilon^{-2}+\epsilon^{-4}\log\frac1\epsilon)\log\frac n\delta)$ comparisons and fewer calls to \OS{}
and \OP{}, so its expected oracle complexity is $\tilde O((n/\epsilon^2+1/\epsilon^4)\log(n/\delta))$. Up to logarithmic factors, the term
$n/\epsilon^2$ dominates once $n\ge\epsilon^{-2}$. For smaller $n$ the term $1/\epsilon^4$ dominates, and for $n\le1/\epsilon$ the uniform scheme's
$O(n^2\epsilon^{-2}\log\frac n\delta)$ is smaller still. Besides the oracle calls, MWU updates $n+1$ weights per round, which adds
$O(n\log(n/\delta)/\epsilon^2)$ arithmetic operations, and drawing indices from the weights costs $O(\log n)$ each with prefix
sums. In a binomial round that uses a sample of $p$, drawing the rank costs $O(\epsilon^{-2}\log\frac1\epsilon)$ operations (Lemma~\ref{lem:binom}).
We count real arithmetic (including logarithms and exponentials), uniform draws and operations on sampled points at unit cost. Compared with the proper factor-$3$ algorithm of~\cite{Zou26}, we get factor $2$
at the price of the term $1/\epsilon^4$, up to logarithmic factors. 

\paragraph{An answer to~\cite[Question~1]{AACS26}.} Our algorithm achieves the factor $2$ on countable domains with
$O(\log(n/\delta)/\epsilon^2)$ samples of $p$. By the previous paragraph its expected running time is $\tilde O((n/\epsilon^2+1/\epsilon^4)\log(n/\delta))$, independent of $\X$.
In the model of~\cite{AACS26} a comparison costs two value queries, so the bound also holds there.
The algorithm averages weighted quantiles of the values $q_i(x)$ and then normalizes. The result can lie outside the convex hull of the $q_i$, so the lower bound for mixtures does not apply to it. Aamand et al.\ also note that it was unclear which classes of outputs beyond mixtures could beat the
factor $3$ with a description of length polynomial in $n$. Our description has $O(n\log(n/\delta)/\epsilon^2)$ real numbers.

\paragraph{Comparison with~\cite{BBEKM21}.} The factor-$2$ learner of Bousquet et al.\ uses
\[
  \tilde O\Bigl(\frac{\log n\cdot\min\{\log n,\log(1/\delta)\}+\log(1/\delta)}{\epsilon^2}\Bigr)
\]
samples of $p$~\cite[Theorem~1]{BBEKM21}. Theorem~\ref{thm:mwu} bounds the number of rounds, and each uses at most one sample of $p$.
Proposition~\ref{prop:normsafe} gives the guarantee for $\bar u$. Our sample bound improves on theirs in two ways. First, it has no hidden logarithmic factors. Second, $\log n$ and $\log(1/\delta)$ add rather than multiply. For $\delta=n^{-c}$ with a
constant $c$, their bound is $\tilde O(\log^2n/\epsilon^2)$ and ours is $O(\log n/\epsilon^2)$. Their learner makes only
statistical queries to $p$, and it needs operations on $\X$ beyond our oracles. Ours uses samples of $p$ directly,
together with \OS{} and \OC{}.

\paragraph{Comparison with~\cite{AACSSZ26}.}\label{par:concurrent} Aamand et al.\ achieve the same optimal statistical guarantee for improper hypothesis selection.
With probability at least $1-\delta$, they give a factor-$2$ approximation up to additive $\epsilon$ using $O(\epsilon^{-2}\log(n/\delta))$ samples from $p$.
Both approaches use multiplicative weights and pointwise weighted quantiles, but the algorithms differ in their constraints and oracle requirements.
Their algorithm maintains two constraints for each candidate~\cite[Eq.~(34)]{AACSSZ26}, while ours maintains one constraint for each candidate and one additional constraint on the total mass.
Moreover, both algorithms draw samples from the candidates. Theirs also evaluates candidate densities, whereas ours only compares them pointwise.
Comparisons are available, for example, when the candidate densities at each point are known up to a common positive factor.

They also obtain optimal results for proper and mixture outputs. On the other hand, for improper selection, we give four implementations with different trade-offs between candidate samples and comparisons.
Our binomial implementation uses comparisons against all $n$ candidates during construction to update the weights.
However, once construction is complete, its sampler no longer needs this full scan.
Consequently, for fixed $0<\epsilon<1$ and cap parameter $\beta=\epsilon$, the expected number of oracle calls per output draw is bounded independently of $n$ (Lemma~\ref{lem:output}),
while the per-draw running time bound of Aamand et al.\ is linear in $n$~\cite[Theorem~1.2]{AACSSZ26}.

\subsection{Our techniques}
\label{sec:techniques}

\paragraph{The relaxation.} For a summable function $v$ on $\X$ write $|v|=\sum_xv(x)$ for its mass. Write $a_+=\max\{a,0\}$. For $v\ge0$ and a
distribution $q$ put
\[
  \TV(v,q)\;=\;\sum_x\bigl(q(x)-v(x)\bigr)_+ ,
\]
the mass of $q$ that $v$ fails to cover. On distributions this is the usual total variation distance. We
solve the relaxed problem
\begin{equation}\label{eq:relaxed}
  v\ge0,\qquad |v|\le1,\qquad \TV(v,q_i)\le\TV(p,q_i)\quad(i=1,\dots,n),
\end{equation}
up to additive $\epsilon$, and then normalize. A mass deficit counts fully against $v$, so normalizing costs
only $O(\epsilon)$ (Proposition~\ref{prop:normsafe}). The other one-sided sum, $\sum_x(v(x)-q(x))_+$, would make
$v=0$ feasible, so it does not give a useful relaxation (Remark~\ref{rem:otherTV}).

\paragraph{Multiplicative weights.} We write~\eqref{eq:relaxed} as $n+1$ convex constraints, one for the mass and one
for each $q_i$, and run multiplicative weights (MWU) over them~\cite{AHK12}. In each round MWU asks for a point $v\ge0$
that does well against the weighted constraints, and then it shifts weight towards the constraints that $v$ violates.
The gains may be random and may be divided by $1+|v|$, a factor we never compute. If they are bounded, have small weighted mean,
and cover the scaled constraints in conditional expectation up to $O(\epsilon)$, then $O(\log(n/\delta)/\epsilon^2)$
rounds suffice (Theorem~\ref{thm:mwu}). When the mass constraint has weight below $\frac12$, the best response is, at
each point, a weighted quantile of the values $q_i(x)$ (Proposition~\ref{prop:best}). Comparisons identify an index at the quantile.
The linear gains of the best response meet MWU's needs with no slack.

\paragraph{The race.} A proposal draws an index $I$ and a point $Y\sim q_I$, and an acceptance test keeps it with some
probability, so that accepted points follow the normalized iterate whenever it has positive mass. We run proposals against a coin that stops
with one sample $X\sim p$. With the full acceptance test, the signed subgradient at the stopping point estimates
the gains divided by $1+|v|$ without bias (Theorem~\ref{thm:race}). This is why a round needs at most one sample of $p$.

\paragraph{Four proposal schemes.} Sections~\ref{sec:uniform}--\ref{sec:smooth} build proposals and acceptance tests
in four ways. The \emph{uniform} scheme runs the exact best response with no bias. It needs at most $n$ proposals per round
in expectation and $n-1$ comparisons per test. The \emph{DKW} scheme works with $m=O(\epsilon^{-2}\log(n/\epsilon))$ indices drawn from
the weights, which replaces $n$ by $m$ almost everywhere. The \emph{rank-smoothed} schemes blur the quantile's rank
over a window of width $\Theta(\epsilon)$, so a race needs only $O(1/\epsilon)$ proposals in expectation. The window kernel tests the rank with a full scan. The binomial kernel tests it with random
comparisons and stops early on clear rejections, so a test costs $\tilde O(1/\epsilon)$ comparisons on average for each choice of the kernel rank.

\paragraph{Organization.} Section~\ref{sec:framework} develops the framework: MWU with scaled gains, the ideal round,
the best response as a quantile, proposal schemes, the race and the algorithm. Sections~\ref{sec:uniform},
\ref{sec:dkw} and~\ref{sec:smooth} give the four proposal schemes. Section~\ref{sec:cost} collects the oracle costs in one
table.

\section{The framework}\label{sec:framework}

This section works from the top down. First we state what MWU needs from each round and prove that it
suffices (\S\ref{sec:mwu}). Then we describe an ideal round, in which linear gains at the best response,
estimated from one sample of $p$, meet MWU's needs with no slack (\S\ref{sec:iterates}). Next we characterize the
best response as a weighted quantile (\S\ref{sec:quantile}). Then we relax the round and say what a proposal scheme
supplies (\S\ref{sec:generic}). We turn a proposal scheme into gain estimates by a race (\S\ref{sec:race}), and assemble the
algorithm (\S\ref{sec:algo}).

The main idea is simple. MWU never needs exact gains. It needs bounded gains whose weighted average is small in expectation
and whose individual means cover the scaled constraints, up to a small bias. Each round's gains may be multiplied by a positive factor, as
long as these factors are not small on average. Bias of order $\epsilon$ costs us only an additive $\epsilon$ in the
final guarantee, and boundedness keeps the number of rounds logarithmic in $n$. This freedom lets us estimate the gains
by sampling. Dividing by $1+|v|$ keeps the estimates bounded whatever the mass
of the response. It also lets us use a response that is not the best one, as long as its divided weighted gain is
small.

\subsection{MWU with scaled gains}\label{sec:mwu}

We encode the constraints as
\begin{equation}\label{eq:fdef}
  f_0(v)=|v|-1,\qquad f_i(v)=\TV(v,q_i)-\TV(p,q_i)\quad(i=1,\dots,n),
\end{equation}
for $v\ge0$ of finite mass. Each $f_i$ is convex on $v\ge0$, since $c\mapsto(q-c)_+$ is convex, and every constraint is tight
at $p$: $f_i(p)=0$ for $i=0,\dots,n$. We call constraint $0$ the \emph{mass expert} and constraints
$1,\dots,n$ the \emph{distance experts}.

MWU keeps a vector $y\in\R^{n+1}$, initialized to all ones. In round $t$ it plays the weights
$w_t=y/\sum_iy_i$, receives a gain vector $\hat g_t\in[-1,1]^{n+1}$, and updates $y_i\leftarrow y_ie^{\eta\hat g_{t,i}}$.
The round also produces an iterate $v_t\ge0$, which may be random. We say that everything that happened
before round $t$ is the \emph{past}, and we put $\lambda_t=1/(1+|v_t|)$. The gains will estimate the linear gains of
$v_t$ multiplied by $\lambda_t$, a factor that the algorithm does not know.

\begin{theorem}[MWU with scaled gains]\label{thm:mwu}
Fix $\epsilon\in(0,1]$ and $\delta\in(0,1)$. Suppose that every round produces an iterate $v_t\ge0$ of finite mass and gains
$\hat g_t\in[-1,1]^{n+1}$ such that
\begin{enumerate}
\item[(S1)]\phantomsection\label{cond:S1} the weighted gain is small on average: $\Ex[\langle w_t,\hat g_t\rangle\mid\text{past}]\le\epsilon/12$;
\item[(S2)]\phantomsection\label{cond:S2} each gain overestimates its scaled constraint, up to $\epsilon/12$:
$\Ex[\hat g_{t,i}\mid\text{past},v_t]\ge\lambda_tf_i(v_t)-\epsilon/12$ for every $i=0,\dots,n$.
\end{enumerate}
By the tower property, it suffices to prove \Stwo{} given the past and all randomness used to choose $v_t$. Run $T=\lceil900\ln(2(n+1)/\delta)/\epsilon^2\rceil$ rounds
with $\eta=\sqrt{\ln(n+1)/T}$. Then with probability at least $1-\delta$ we have $\sum_t\lambda_t\ge T/3$, and the weighted
average
\[
  \bar v\;=\;\frac{\sum_t\lambda_tv_t}{\sum_t\lambda_t}
\]
satisfies $\bar v\ge0$ and $f_i(\bar v)\le\epsilon$ for every $i=0,\dots,n$.
\end{theorem}

\begin{proof}
We combine the regret bound of MWU with two martingale bounds, and then use the mass constraint. Write
$\Lambda_\delta=\ln(2(n+1)/\delta)$.

\emph{Regret.} Put $\Phi_t=\sum_iy_i$ at the start of round $t$, so $\Phi_1=n+1$. Since $|\eta\hat g_{t,i}|\le1$, we have
$e^{\eta\hat g_{t,i}}\le1+\eta\hat g_{t,i}+\eta^2$, hence $\Phi_{t+1}/\Phi_t\le1+\eta\langle w_t,\hat g_t\rangle+\eta^2$. Taking logarithms and
using $\ln(1+a)\le a$ gives $\ln\Phi_{T+1}\le\ln(n+1)+\eta\sum_t\langle w_t,\hat g_t\rangle+\eta^2T$. Also
$\ln\Phi_{T+1}\ge\eta\sum_t\hat g_{t,j}$ for each $j$. With our choice of $\eta$,
\[
  \sum_t\hat g_{t,j}\;\le\;\sum_t\langle w_t,\hat g_t\rangle+2\sqrt{T\ln(n+1)}.
\]

\emph{Martingales.} The differences $\Ex[\hat g_{t,j}\mid\text{past},v_t]-\hat g_{t,j}$ have mean zero given the past and
$v_t$, and under this conditioning each lies in an interval of length $2$. Hoeffding's lemma bounds its conditional
exponential moment at $a>0$ by $e^{a^2/2}$. Averaging over $v_t$ gives the same bound given the past. The usual
martingale bound and a union bound over $j$ show
that, with probability at least $1-\delta/2$, their sum is at most $\sqrt{2T\Lambda_\delta}$ for every $j$. The same
argument applies to $\langle w_t,\hat g_t\rangle-\Ex[\langle w_t,\hat g_t\rangle\mid\text{past}]$, with probability at least $1-\delta/2$.

\emph{Putting it together.} On both events, \Stwo{}, the regret bound and \Sone{} give, for every $j$,
\[
  \sum_t\lambda_tf_j(v_t)\;\le\;\frac{\epsilon T}{12}+\sum_t\Ex[\hat g_{t,j}\mid\text{past},v_t]
  \;\le\;\frac{\epsilon T}6+5\sqrt{T\Lambda_\delta}\;\le\;\frac{\epsilon T}3\;\eqqcolon\;B,
\]
where the last step uses the value of $T$.

\emph{The mass constraint.} Take $j=0$. Since $f_0(v)=|v|-1$, we have $\lambda_tf_0(v_t)=(|v_t|-1)/(|v_t|+1)=1-2\lambda_t$.
So $T-2\sum_t\lambda_t\le B$, and $\sum_t\lambda_t\ge(T-B)/2\ge T/3$. Each $f_j$ is convex, and $\bar v$ is a convex combination of
the $v_t$, so Jensen's inequality gives
\[
  f_j(\bar v)\;\le\;\frac{\sum_t\lambda_tf_j(v_t)}{\sum_t\lambda_t}\;\le\;\frac{2B}{T-B}\;=\;\frac{2\epsilon}{3-\epsilon}\;\le\;\epsilon .
\]
Finally $\bar v\ge0$ because every $v_t\ge0$.
\end{proof}

The rest of the framework builds rounds that satisfy \Sone{} and \Stwo{}.

\subsection{The ideal round}\label{sec:iterates}

We first describe an ideal round, in which we could compute everything exactly. MWU's best response
minimizes the weighted constraints over all $v\ge0$. Its mass can grow with $n$, but the scaled gains do not
need a bound on the mass. The best response meets the conditions of Theorem~\ref{thm:mwu} with no slack, by convexity and
minimality alone. One sample of $p$ estimates its scaled linear gains without bias, and the estimate is bounded by
$1$. The oracles do not let us compute it, however. Sections~\ref{sec:quantile}--\ref{sec:race} take up that problem.

\paragraph{The best response.} Fix the weights $w$ of a round, and let $u^\ast$ be a minimizer of $\sum_{i=0}^nw_if_i$
over all $v\ge0$. Proposition~\ref{prop:best} below exhibits one, of finite mass. We take subgradients on $\ell_1(\X)$, using
$f_0(v)=\sum_xv(x)-1$ and $f_i(v)=\sum_x(q_i(x)-v(x))_+-\TV(p,q_i)$ for $i\ge1$.
These functions are convex and $1$-Lipschitz, so every such subgradient $G$ has $|G(x)|\le1$ at every point.
For $h\in\ell_\infty(\X)$ and $z\in\ell_1(\X)$ write $\langle h,z\rangle=\sum_x h(x)z(x)$.

\begin{lemma}[First-order optimality]\label{lem:foc}
There are subgradients $G_i$ of $f_i$ at $u^\ast$, for $i=0,\dots,n$, such that
\[
  \sum_{i=0}^nw_i\,\langle G_i,\,u-u^\ast\rangle\;\ge\;0\qquad\text{for every } u\ge0 \text{ of finite mass}.
\]
\end{lemma}

\begin{proof}
The point $u^\ast$ minimizes $\sum_iw_if_i$ plus the indicator of the cone $\{v\ge0\}$. Each $f_i$ is finite,
continuous and convex on $\ell_1(\X)$, so the Moreau--Rockafellar sum rule applies. Zero is the sum of an element
$\sum_iw_iG_i$ of $\sum_iw_i\,\partial f_i(u^\ast)$ and a normal vector $N$ of the cone at $u^\ast$. A normal vector satisfies
$\langle N,u-u^\ast\rangle\le0$ for every $u$ in the cone.
\end{proof}

Take $G_0,\dots,G_n$ as in Lemma~\ref{lem:foc} and define the \emph{linear gains}
\begin{equation}\label{eq:lingain}
  \ell_i\;=\;\bigl\langle G_i,\,u^\ast-p\bigr\rangle\qquad(i=0,\dots,n).
\end{equation}

\begin{proposition}[Linear gains have no slack]\label{prop:noslack}
Take the iterate $v_t=u^\ast$ and the gains \eqref{eq:lingain}. Then $\langle w,\ell\rangle\le0$, and
$\ell_i\ge f_i(u^\ast)$ for every $i$.
\end{proposition}

\begin{proof}
Lemma~\ref{lem:foc} with $u=p$ gives $\langle w,\ell\rangle\le0$. For each $i$, convexity gives
$f_i(p)\ge f_i(u^\ast)+\langle G_i,p-u^\ast\rangle$. Every constraint is tight at $p$, so $f_i(p)=0$.
Rearranging gives $\ell_i\ge f_i(u^\ast)$.
\end{proof}

\paragraph{One sample from $p$ suffices.} The linear gains still involve $p$. The term $\langle G_i,p\rangle$ is
the mean of $G_i(X)$ for $X\sim p$. So one sample from $p$ estimates it without bias, for all $n+1$ experts at
once. Dividing by $1+|u^\ast|$ makes the estimate bounded.

\begin{proposition}[Scaled gains from one sample of $p$]\label{prop:onesample}
Suppose we know the function $G_i$ and the number $\langle G_i,u^\ast\rangle$ for every $i$, and put
$\lambda=1/(1+|u^\ast|)$. Draw $X\sim p$ and set $\hat g_i=\lambda(\langle G_i,u^\ast\rangle-G_i(X))$. Then $\Ex\hat g_i=\lambda\ell_i$ and
$|\hat g_i|\le1$. So \Sone{} and \Stwo{} hold with no slack.
\end{proposition}

\begin{proof}
The mean of $G_i(X)$ is $\langle G_i,p\rangle$. The iterate is fixed by the past, so conditioning on it changes
nothing. Then $\Ex\langle w,\hat g\rangle=\lambda\langle w,\ell\rangle\le0$ and $\Ex\hat g_i=\lambda\ell_i\ge\lambda f_i(u^\ast)$, by
Proposition~\ref{prop:noslack}. The bound uses $|G_i|\le1$: $|\hat g_i|\le\lambda(|u^\ast|+1)=1$.
\end{proof}

\begin{corollary}[The ideal algorithm]\label{cor:ideal}
MWU with these gains and $T=\lceil900\ln(2(n+1)/\delta)/\epsilon^2\rceil$ rounds returns, with probability at least
$1-\delta$, a weighted average $\bar v\ge0$ with $f_i(\bar v)\le\epsilon$ for all $i$. It uses one sample from $p$ per round.
\end{corollary}

\paragraph{Extension to other distances.} Theorem~\ref{thm:mwu} and Corollary~\ref{cor:ideal} use very few properties of total
variation. The proof of Theorem~\ref{thm:mwu} uses only that each $f_i$ is convex on $v\ge0$, in its last step, and that
the mass expert is $f_0(v)=|v|-1$. 
Lemma~\ref{lem:foc} and
Proposition~\ref{prop:noslack} use convexity, a minimizer $u^\ast$, tightness at $p$, and the extension of each $f_i$ to
$\ell_1(\X)$. Proposition~\ref{prop:onesample} uses $|G_i(x)|\le1$ to keep the scaled gains in $[-1,1]$. So let $d(v,q)$
be any function of $v\ge0$ that is convex and $1$-Lipschitz in the $\ell_1$ norm, and put $f_i(v)=d(v,q_i)-d(p,q_i)$.
Such a function extends to a convex $1$-Lipschitz function on $\ell_1(\X)$, for example
$\inf_{u\ge0}\bigl(d(u,q)+\|v-u\|_1\bigr)$, and the subgradients of the extension have $\ell_\infty$ norm at most $1$. So
Lemma~\ref{lem:foc} and Proposition~\ref{prop:onesample} still hold. If the weighted constraints have a
minimizer over $v\ge0$, the ideal algorithm returns $\bar v\ge0$ with $|\bar v|\le1+\epsilon$ and $d(\bar v,q_i)\le d(p,q_i)+\epsilon$ for
every $i$, with the same $T$ and one sample of $p$ per round. What is specific to total variation is the rest of the
paper: the best response as a quantile (\S\ref{sec:quantile}), which lets comparisons replace the exact subgradients,
and the normalization of $\bar v$ (Proposition~\ref{prop:normsafe}).


\subsection{The best response as a quantile}\label{sec:quantile}

The ideal round needs the subgradients $G_i$ and the numbers $\langle G_i,u^\ast\rangle$. Neither is available,
since each number is a sum over all of $\X$. So we replace them by randomized estimates, and for that we need
an explicit description of the best response. The objective splits over points. At each point the mass
expert pulls the value down with weight $w_0$, and each distance expert whose value lies above pulls it up
with its weight. The balance point is a weighted quantile of the $q_i(x)$, at a rank that $w_0$ determines.
Its subgradients are simple functions of comparisons.

\paragraph{Quantiles and subgradients at a rank.} Let $\tilde w$ be a probability vector on $[n]$. Fix a point
$x$ and arrange the indices in \emph{groups}: the indices with a common value $q_i(x)$ form one group, and the
groups are ordered by that value. Fix a rank $s\in(0,1)$. Walk through the groups in order and accumulate
their $\tilde w$-weights. Let $Q_s(x)$ be the value of the group at which the running total first reaches $s$.
This is the lower weighted $s$-quantile. Write $Q_s^{\tilde w}$ when the weights need to be shown. Let $W_A$ be the weight of the earlier groups and $W_T$ the weight of
this group. By the choice of the group,
\begin{equation}\label{eq:sandwich}
  W_A\;<\;s\;\le\;W_A+W_T,
\end{equation}
so $W_T>0$. For $i=1,\dots,n$ define the comparison sign $g_i(x,s)$ and the subgradient $d_i(x,s)$ by
\begin{equation}\label{eq:gdef}
  g_i(x,s)=
  \begin{cases}
    +1 & \text{if } q_i(x)<Q_s(x),\\
    \gamma_s(x)\coloneqq\dfrac{2s-2W_A-W_T}{W_T} & \text{if } q_i(x)=Q_s(x),\\
    -1 & \text{if } q_i(x)>Q_s(x),
  \end{cases}
  \qquad
  d_i(x,s)\;=\;\tfrac12\bigl(g_i(x,s)-1\bigr).
\end{equation}
Write $Q_s$, $g_i(\cdot,s)$ and $d_i(\cdot,s)$ for the functions of $x$, and $\tilde q=\sum_i\tilde w_iq_i$.

\begin{lemma}[Facts about quantiles at a rank]\label{lem:facts}
For every $x$ and $s\in(0,1)$:
\begin{enumerate}
\item[(a)] $\gamma_s(x)\in(-1,1]$, so $d_i(x,s)\in[-1,0]$, and $d_i(\cdot,s)$ is a subgradient of $f_i$ at $Q_s$ for
each $i\ge1$; the constant $1$ is a subgradient of $f_0$ at every $v\ge0$;
\item[(b)] $\sum_i\tilde w_i\,g_i(x,s)=2s-1$;
\item[(c)] $Q_s(x)\le\tilde q(x)/(1-s)$; in particular $|Q_s|\le1/(1-s)$;
\item[(d)] $s\mapsto Q_s(x)$ is nondecreasing, and $\int_0^1Q_s(x)\,ds=\tilde q(x)$.
\end{enumerate}
\end{lemma}

\begin{proof}
(a) By \eqref{eq:sandwich}, $2s-2W_A-W_T$ lies in $(-W_T,W_T]$. The subdifferential of $c\mapsto(q-c)_+$ is
$\{-1\}$ for $c<q$, $\{0\}$ for $c>q$ and $[-1,0]$ at $c=q$. The function $d_i$ takes these values, and summing
over $x$ gives a subgradient of $f_i$. The mass constraint has gradient $1$.
(b) The weighted sum is $W_A-(1-W_A-W_T)+W_T\gamma_s(x)$, which simplifies to $2s-1$.
(c) The groups from that of $Q_s(x)$ onwards carry weight at least $1-s$, and their values are at least
$Q_s(x)$. So $\tilde q(x)\ge(1-s)Q_s(x)$. Sum over $x$ and use $|\tilde q|=1$.
(d) Each group's value is taken on a rank interval of length equal to its weight, in increasing order of
value.
\end{proof}

\paragraph{The best-response rank.} The MWU weights are positive. Write $D=1-w_0=\sum_{i\ge1}w_i$, and let $\bar w_i=w_i/D$ for $i=1,\dots,n$. Define
\begin{equation}\label{eq:sstar}
  s^\ast\;=\;1-\frac{w_0}{D}\;=\;\frac{1-2w_0}{1-w_0}.
\end{equation}

\paragraph{The weighted subgradient density.} At the quantile $Q_s$, the weighted objective
$F_w=\sum_{i=0}^nw_if_i$ has the subgradient $\sum_iw_iG_i$ with $G_0=1$ and $G_i=d_i(\cdot,s)$, by
Lemma~\ref{lem:facts}(a). We call its value at a point,
\[
  \psi_s(x)\;=\;w_0\cdot1+\sum_{i\ge1}w_i\,d_i(x,s),
\]
the \emph{weighted subgradient density}. It is a subgradient, at $c=Q_s(x)$, of the one-point objective
$c\mapsto w_0c+\sum_{i\ge1}w_i(q_i(x)-c)_+$ into which $F_w$ splits. At $Q_s$ the linear gains are $\ell_i=\langle G_i,Q_s-p\rangle$, so
$\langle w,\ell\rangle=\sum_x\psi_s(x)\bigl(Q_s(x)-p(x)\bigr)$. The next lemma shows that, under
the true weights, the density does not depend on the point at all.

\begin{lemma}[The weighted subgradient density]\label{lem:density}
Take $\tilde w=\bar w$. For every $x$ and $s\in(0,1)$, $\psi_s(x)=D\,(s-s^\ast)$. When $w_0<\frac12$, the density is negative below the
rank $s^\ast$, zero at it, and positive above it.
\end{lemma}

\begin{proof}
By Lemma~\ref{lem:facts}(b), $\sum_{i\ge1}w_id_i=\frac D2\sum_i\bar w_ig_i-\frac D2=\frac D2(2s-1)-\frac D2=D(s-1)$. Add
$w_0=D(1-s^\ast)$, which is the definition \eqref{eq:sstar}.
\end{proof}

\begin{proposition}[The best response]\label{prop:best}
If $w_0<\frac12$, then $Q_{s^\ast}$ under $\bar w$ minimizes $\sum_{i=0}^nw_if_i$ over all $v\ge0$. Its weighted
subgradient density is $\psi_{s^\ast}=0$ at every point, and $|Q_{s^\ast}|\le(1-w_0)/w_0$. If $w_0\ge\frac12$, the zero function is a
minimizer.
\end{proposition}

\begin{proof}
The objective is $\sum_x\bigl(w_0v(x)+\sum_{i\ge1}w_i(q_i(x)-v(x))_+\bigr)$ minus a constant. At $x$, the right
slope of $c\mapsto w_0c+\sum_{i=1}^nw_i(q_i(x)-c)_+$ is $w_0-D\,\bar w\{i:q_i(x)>c\}$. If $w_0\ge\frac12$, then $w_0/D\ge1$ and the slope is nonnegative, so $0$ is a minimizer.
Now suppose $w_0<\frac12$. The slope is negative while the experts
above $c$ carry $\bar w$-weight more than $w_0/D=1-s^\ast$, and nonnegative once they carry at most that. So
$Q_{s^\ast}(x)$ is a minimizer. The density
is Lemma~\ref{lem:density} at $s=s^\ast$. The mass bound is Lemma~\ref{lem:facts}(c).
\end{proof}

\subsection{Relaxed rounds and proposal schemes}\label{sec:generic}

We now know what to estimate. The subgradients at the sample $X$ are cheap: comparisons at $X$ locate the
quantile and classify every index (Lemma~\ref{lem:cheap}). The numbers $\langle d_i(\cdot,s),Q_s\rangle$ are not,
since each is a sum over all of $\X$. We estimate them by sampling. Every proposal scheme produces random points whose law
is the iterate itself, up to a known constant, through a proposal--acceptance scheme, as in rejection sampling. To
make the sampling cheap we relax the round in two ways. The rank may be random, drawn from a \emph{rank kernel} $\pi$,
instead of fixed. The quantiles may be taken under a \emph{reference vector} $\tilde w$ that approximates $\bar w$. This
subsection defines the relaxed rounds and the proposal schemes, and says what a proposal scheme must guarantee.
Section~\ref{sec:race} turns a proposal scheme into gain estimates.

\paragraph{Relaxed iterates and their gains.} Let $\tilde w$ be a probability vector on $[n]$ and $\pi$ a
probability measure on ranks. Averaging the quantiles and their linear gains over the rank gives
\begin{equation}\label{eq:iterate}
  v=\int Q_s\,\pi(ds),\qquad \ell_0=|v|-1,\qquad
  \ell_i=\int\bigl\langle d_i(\cdot,s),\,Q_s-p\bigr\rangle\,\pi(ds)\quad(i\ge1),
\end{equation}
with $Q_s$ and $d_i$ taken under $\tilde w$. The integrals are finite since $Q_s(x)\le\sum_{i=1}^n q_i(x)$ and $|d_i(x,s)|\le1$. The error of the reference vector enters through the function
\[
  \eta_s(x)\;=\;\sum_{i\ge1}(\bar w_i-\tilde w_i)\,g_i(x,s),
\]
which vanishes when $\tilde w=\bar w$.

\paragraph{Kernels and their ranks.} Each proposal scheme supplies a family of kernels $\pi_s$, one for every
rank $s\in(0,1)$. Each $\pi_s$ is a probability measure on $(0,1)$. A \emph{point rank kernel} has $\pi_s=\delta_s$, the point mass at $s$. A \emph{window kernel} of bandwidth $\tilde\epsilon>0$
has $\pi_s$ uniform on $[s-\frac{\tilde\epsilon}2,s+\frac{\tilde\epsilon}2]\cap(0,1)$, the window cut off at $0$ and $1$. Write
$v_s=\int Q_t\,\pi_s(dt)$ for the iterate at rank $s$, so that $|v_s|$ is its mass.

\begin{definition}[Proposal scheme]\label{def:scheme}
Given the weights of a round with $w_0<\frac12$, a \emph{proposal scheme} fixes a reference vector $\tilde w$, a family of kernels
$\pi_s$ and, for each $s$, a known rejection constant $\Gamma=\Gamma_s\ge1$. These may depend on randomness drawn at the
start of the round. Given this choice, the scheme can draw from $\pi_s$ without an oracle. It also supplies a
\emph{proposal routine}. Each call uses fresh randomness to draw an index $I$ and a point $Y\sim q_I$ with one call to \OS{}.
The proposal has a rank $r$ and uses comparisons and internal randomness to decide whether to accept. We require three things.
\begin{enumerate}
\item[(a)] The \emph{proposal property}: given the choice made at the start of the round, for every $x\in\X$ and every Borel set
$E\subseteq(0,1)$,
\begin{equation}\label{eq:proposal}
  \Pr\bigl[Y=x,\ r\in E,\ \text{accepted}\bigr]\;=\;\frac1\Gamma\int_EQ_t(x)\,\pi_s(dt).
\end{equation}
\item[(b)] On acceptance, $I$ lies in the group of $Q_r(Y)$, and the routine supplies $r$ or its tie value $\gamma_r(Y)$.
\item[(c)] The \emph{bias condition}: let the kernel $\pi_{s^\ast}$ at the best-response rank have iterate $v$ and gains $\ell$ as in
\eqref{eq:iterate}, and put $\lambda=1/(1+|v|)$. Then
\begin{equation}\label{eq:bias}
  \Ex\bigl[\lambda\langle w,\ell\rangle\bigr]\;\le\;\frac\epsilon{12},
\end{equation}
where the expectation is over the choice made at the start of the round.
\end{enumerate}
\end{definition}

Summing \eqref{eq:proposal} over $x$ shows that a proposal at rank $s$ is accepted with probability $|v_s|/\Gamma$.

\begin{lemma}[Linear gains of a relaxed round]\label{lem:wgain}
Let $v$ and $\ell$ be as in \eqref{eq:iterate}.
\begin{enumerate}
\item[(i)] $v\ge0$, $f_0(v)=\ell_0$, and $f_i(v)\le\ell_i$ for $i\ge1$.
\item[(ii)] The weighted gain is
\begin{equation}\label{eq:wgain}
  \langle w,\ell\rangle\;=\;D\int\Bigl\langle (s-s^\ast)+\tfrac12\eta_s,\;Q_s-p\Bigr\rangle\,\pi(ds).
\end{equation}
\end{enumerate}
\end{lemma}

\begin{proof}
(i) For $i\ge1$, Lemma~\ref{lem:facts}(a) and convexity give $f_i(Q_s)\le\langle d_i(\cdot,s),Q_s-p\rangle$. Average over
$\pi$, and use convexity of $f_i$ to pass to $v$.

(ii) The weighted gain is $\langle w,\ell\rangle=w_0\ell_0+\sum_{i\ge1}w_i\ell_i$. The mass expert's gain is itself a linear
gain, the one of its subgradient $1$. Since $|p|=1$ and $|v|=\int|Q_s|\,\pi(ds)$,
\[
  \ell_0\;=\;|v|-1\;=\;\int\bigl\langle 1,\,Q_s-p\bigr\rangle\,\pi(ds).
\]
So the term $w_0\ell_0$ joins the others inside one integral:
$\langle w,\ell\rangle=\int\langle\psi_s,Q_s-p\rangle\,\pi(ds)$, with the weighted subgradient density
$\psi_s=w_0+\sum_{i\ge1}w_id_i(\cdot,s)$ taken with the subgradients under $\tilde w$. By Lemma~\ref{lem:facts}(b) for $\tilde w$, $\sum_i\bar w_ig_i=2s-1+\eta_s$. So
$\psi_s=w_0+\frac D2(2s-1+\eta_s)-\frac D2=D(s-s^\ast)+\frac D2\eta_s$, as in the proof of Lemma~\ref{lem:density}. This is
\eqref{eq:wgain}.
\end{proof}

Part~(i) says that the linear gains of any relaxed round overestimate the constraints. Part~(ii) is the tool for the
bias condition. The bracket in \eqref{eq:wgain} has two parts. The first, $s-s^\ast$, is the distance of a rank in the
kernel from the best-response rank. The second, $\frac12\eta_s$, is the error of the reference vector. Each proposal
scheme makes one of them vanish and the other small. The factor $\lambda$ in \eqref{eq:bias} is needed because $\|Q_s-p\|_1$ can be as large as $|Q_s|+1$, and $|Q_s|$ can grow with $n$.

\paragraph{Cheap subgradients.} A race needs the subgradients at its stopping point, which is either an accepted
point or the sample $X$. Draws from
$\bar w$ or $\tilde w$, uniform variables and arithmetic on weights use no oracle.

\begin{lemma}[Cheap subgradients]\label{lem:cheap}
Fix $x$ and $s$. If we know an index $j$ in the group of $Q_s(x)$, then $2(n-1)$ comparisons give $g_i(x,s)$ and
$d_i(x,s)$ for all $i$. Otherwise a weighted quickselect over the support of $\tilde w$ finds such an index with
$O(k)$ comparisons, where $k$ is the size of the support.
\end{lemma}

\begin{proof}
Ask both $q_i(x)>q_j(x)$ and $q_j(x)>q_i(x)$ for every $i\ne j$. The answers place every $i$ below, tied with, or
above $Q_s(x)$, and the known weights give $W_A$, $W_T$ and $\gamma_s(x)$. To find $j$, run quickselect with
three-way partitions, recursing into the part whose cumulative weight contains $s$. Use median-of-medians to find the
median by count at each step. The partition costs linear time and at most half the indices remain.
\end{proof}

At an accepted point the proposed index $I$ serves as $j$. With the supplied rank or tie value, $2(n-1)$ comparisons give every subgradient. At $X$
we first run quickselect, at cost $O(n)$, or $O(k)$ plus a sweep of all $n$ indices when $\tilde w$ has support of
size $k$. The race of the next subsection needs subgradients at only one of these points per round.

\subsection{The race}\label{sec:race}

When $|v_s|>0$, a proposal scheme gives samples from the normalized iterate by running proposals until one is accepted.
The accepted pair $(Y,r)$ has law proportional to $Q_t(x)\,\pi_s(dt)$, and $d_i(Y,r)$ estimates $\int\langle d_i(\cdot,t),Q_t\rangle\,\pi_s(dt)/|v_s|$
without bias. This alone does not give the gains, because they need the scale $|v_s|$. An unbiased estimate of
$|v_s|$ built from acceptance events can be as large as $\Gamma$. The time to the first acceptance, $\Gamma/|v_s|$ proposals
on average, is unbounded when the mass is small. With $k$ equally weighted candidates on disjoint supports and $s^\ast\le1-1/k$, for example, the best
response is the zero function. We fix both problems by padding the iterate with $p$ and dividing the gains by
$1+|v_s|$. The padded measure has mass at least $1$, so the race below ends after at most $\Gamma$ proposals on
average. The divided gains are linear in the law of the stopping point, so a single signed subgradient estimates
them.

In this subsection we fix the round's setup and the rank $s$, with iterate $v=v_s$ and linear gains $\ell$ as in
\eqref{eq:iterate}, and we put $\lambda=1/(1+|v|)$.

\begin{definition}[The race]\label{def:race}
Repeat the following trial until it stops. Flip a fair coin. On heads, run one proposal at rank $s$, and if it is
accepted, stop with the outcome $(\mathrm V;Y,r)$, where $r$ is the proposal's rank even when the routine supplies only its tie value.
On tails, stop with probability $1/\Gamma$, with the outcome $\mathrm P$. Then output
\[
  \hat g_0=1,\quad \hat g_i=d_i(Y,r)\qquad\text{on }(\mathrm V;Y,r);\qquad\qquad
  \hat g_0=-1,\quad \hat g_i=-d_i(X,t)\qquad\text{on }\mathrm P,
\]
where on $\mathrm P$ we draw $X\sim p$ with one call to \OP{} and independently draw a rank $t\sim\pi_s$ internally.
\end{definition}

The stopping coin on tails uses no oracle. The race is rejection sampling from the
measure $Q_t(x)\,\pi_s(dt)+p(x)\,\pi_s(dt)$ on pairs $(x,t)$, of mass $|v|+1$. The sign records which part the sample came
from.

\begin{lemma}[Law of the race]\label{lem:racelaw}
The outcome satisfies, for every $x\in\X$ and Borel set $E$,
\[
  \Pr\bigl[\mathrm V,\ Y=x,\ r\in E\bigr]=\lambda\int_EQ_t(x)\,\pi_s(dt),\qquad \Pr[\mathrm P]=\lambda .
\]
The expected number of proposals is $\lambda\Gamma\le\Gamma$, and the race calls \OP{} at most once.
\end{lemma}

\begin{proof}
The trials are independent and identically distributed. A trial stops with $(\mathrm V;Y=x,r\in E)$ with probability
$\frac1{2\Gamma}\int_EQ_t(x)\,\pi_s(dt)$, by \eqref{eq:proposal}, and with $\mathrm P$ with probability $\frac1{2\Gamma}$. A sequence of
such trials, stopped at the first one that stops, ends with the law of one trial conditioned on stopping. A trial
stops with probability $(|v|+1)/(2\Gamma)$, so the number of trials is geometric with mean $2\Gamma\lambda$. Each trial is a
proposal with probability $\frac12$, independently of the past, so Wald's identity gives the expected number of
proposals.
\end{proof}

\begin{theorem}[The race estimates the scaled gains]\label{thm:race}
The race outputs $\hat g\in[-1,1]^{n+1}$ with $\Ex\hat g_i=\lambda\ell_i$ for $i=0,\dots,n$. In particular
$\Ex\hat g_i\ge\lambda f_i(v)$ for every $i$, and $\Ex\langle w,\hat g\rangle=\lambda\langle w,\ell\rangle$. The gains use subgradients only at the
stopping point.
\end{theorem}

\begin{proof}
The bound $|\hat g_i|\le1$ holds because $d_i\in[-1,0]$. By Lemma~\ref{lem:racelaw}, $\Ex\hat g_0=\lambda|v|-\lambda=\lambda\ell_0$. For
$i\ge1$, the $\mathrm V$ branch contributes $\lambda\int\langle d_i(\cdot,t),Q_t\rangle\,\pi_s(dt)$. In the $\mathrm P$ branch $X$ and $t$ are
independent, so it contributes $-\lambda\int\langle d_i(\cdot,t),p\rangle\,\pi_s(dt)$. Their sum is $\lambda\ell_i$. The inequality follows from
Lemma~\ref{lem:wgain}(i).
\end{proof}

So the race satisfies \Stwo{} of Theorem~\ref{thm:mwu} with no slack, for any reference vector. At $s=s^\ast$, condition \Sone{}
reduces to a bound on the scaled weighted gain $\Ex[\lambda\langle w,\ell\rangle]$, which is the bias condition of
Definition~\ref{def:scheme}.

\subsection{The algorithm}\label{sec:algo}

\begin{figure}[tbp]
\noindent\rule{\textwidth}{0.6pt}\\[2pt]
\refstepcounter{algorithm}\label{alg:mwu}\textbf{Algorithm \thealgorithm} (MWU with the race)\\[-6pt]
\rule{\textwidth}{0.4pt}\\[2pt]
Set $T=\lceil900\ln(2(n+1)/\delta)/\epsilon^2\rceil$ and $\eta=\sqrt{\ln(n+1)/T}$. Initialize $y=\mathbf 1\in\R^{n+1}$.\\[2pt]
For $t=1,\dots,T$:
\begin{enumerate}[topsep=2pt,partopsep=0pt,itemsep=1pt]
\item Set $w=y/\sum_iy_i$.
\item If $w_0\ge\frac12$, set $\hat g_0=-1$ and $\hat g_i=1$ for $i\ge1$. The iterate is $v_t=0$.
\item Otherwise compute $s^\ast=1-w_0/(1-w_0)$, let the scheme fix $\tilde w$ and its kernel at $s^\ast$, and run the
race of Definition~\ref{def:race}. For binomial proposals use the sequential test, with $v_t=v'$ as in Proposition~\ref{prop:binscheme}(b).
\item Update $y_i\leftarrow y_ie^{\eta\hat g_i}$ for every $i$.
\end{enumerate}
\noindent Return the weights of every round and any other data needed to repeat its setup (the empirical weights for DKW). They determine
$\bar v=\sum_t\lambda_tv_t/\sum_t\lambda_t$, and Lemma~\ref{lem:output} turns them into a sampler.\\[-4pt]
\rule{\textwidth}{0.6pt}
\end{figure}

Algorithm~\ref{alg:mwu} puts the pieces together. It runs MWU, and in every round with $w_0<\frac12$ it runs one race for the
kernel at the best-response rank.

\paragraph{Degenerate rounds.} When $w_0\ge\frac12$ the zero iterate is a best response, by
Proposition~\ref{prop:best}. We take $v_t=0$, so $\lambda_t=1$, and the gains in the degenerate step are the linear gains of the
subgradients $1$ for $f_0$ and $-1$ for every $f_i$ at $0$. These are valid, since $-1$ lies in the subdifferential of
$c\mapsto(q-c)_+$ at $c=0\le q$. They satisfy \Stwo{} exactly: $\hat g_0=-1=f_0(0)$ and $\hat g_i=1\ge1-\TV(p,q_i)=f_i(0)$. The
weighted gain is $-w_0+D=1-2w_0\le0$, so \Sone{} holds. The round uses no oracle.

\begin{proposition}[Normalization is safe]\label{prop:normsafe}
Let $v\ge0$ have mass $m$, with $m\le1+\epsilon$ and $f_i(v)\le\epsilon$ for all $i\ge1$. If $0<m\le1$, the distribution
$\bar u=v/m$ satisfies $\TV(\bar u,q_i)\le\TV(p,q_i)+\epsilon$. If $m>1$, it satisfies
$\TV(\bar u,q_i)\le\TV(p,q_i)+2\epsilon$. If $m=0$, every distribution $u$ satisfies $\TV(u,q_i)\le\TV(p,q_i)+\epsilon$.
\end{proposition}

\begin{proof}
For $v\ge0$, $\sum_x(q-v)_+=\frac12(\|v-q\|_1+1-|v|)$, since $y_+=\frac12(|y|+y)$. So $f_i(v)\le\epsilon$ reads
$\|v-q_i\|_1\le\|p-q_i\|_1+2\epsilon-(1-m)$. For $m>0$, $\|\bar u-v\|_1=|1-m|$. For $0<m\le1$, the triangle inequality gives
$\|\bar u-q_i\|_1\le\|p-q_i\|_1+2\epsilon$. For $m>1$ the two mass terms add up to $2(m-1)\le2\epsilon$. Halve. If $m=0$, then
$1-\TV(p,q_i)\le\epsilon$, and $\TV(u,q_i)\le1$.
\end{proof}

\begin{remark}[The other one-sided extension]\label{rem:otherTV}
With $\sum_x(v(x)-q_i(x))_+$ in place of $\TV(v,q_i)$, the zero function satisfies every constraint, and MWU's best
response is $0$ in every round. That relaxation forgets that the output must become a distribution, so it gives
nothing.
\end{remark}

\paragraph{The output.} We never know the $\lambda_t$, and we never write $\bar v$ down. When $|\bar v|>0$, its normalization
$\bar u=\bar v/|\bar v|$ is proportional to $\sum_t\lambda_tv_t$. If $\bar v=0$, take $\bar u=q_1$. A race for round $t$ ends with $\mathrm V$ with probability
$\lambda_t|v_t|$, and then returns a point with law $v_t/|v_t|$. So choosing a round uniformly and keeping only
$\mathrm V$ outcomes gives point probabilities proportional to $\sum_t\lambda_tv_t(x)$.

\begin{lemma}[Sampling the output]\label{lem:output}
Store each round's weights and setup as in Algorithm~\ref{alg:mwu}. To sample, pick $t$ uniformly from $[T]$ and run a fresh race for
round $t$, using the same acceptance test and treating a degenerate round as an outcome $\mathrm P$. On $\mathrm V$, output $Y$. On $\mathrm P$, start again
without drawing $X$ or a rank. If $|\bar v|>0$, the output has law $\bar u$. On the event of Theorem~\ref{thm:mwu}, the
expected number of attempts is then at most $3/|\bar v|$. On the same event, for any $\beta\in(0,1)$, if we stop after
$K=\lceil(3/\epsilon)\ln(1/\beta)\rceil$ attempts and then output a draw from $q_1$ on failure, the output law $\mu$
satisfies $\TV(\mu,q_i)\le\TV(p,q_i)+2\epsilon+\beta$ for every $i$, including when $\bar v=0$.
\end{lemma}

\begin{proof}
Condition on the run of Algorithm~\ref{alg:mwu}, so that the weights, the iterates $v_t$ and the factors $\lambda_t$ are fixed. The
sampler uses fresh randomness. One attempt picks $t$ uniformly and runs a race for round $t$. By
Lemma~\ref{lem:racelaw}, it outputs $x$ with probability $\frac1T\sum_t\lambda_tv_t(x)$. A degenerate round has $v_t=0$ and never
outputs, so the formula covers it too. This is proportional to $\bar v(x)$. So one attempt succeeds with probability
\[
  a\;=\;\frac1T\sum_t\lambda_t|v_t|\;=\;|\bar v|\,\frac1T\sum_t\lambda_t.
\]
If $\bar v=0$, the capped sampler outputs $q_1$ and Proposition~\ref{prop:normsafe} gives the claim on the stated event.
Otherwise $a>0$, and given success its output has law $\bar u$. The attempts are independent and identically distributed. So the
output of the first successful attempt has law $\bar u$, whatever its index. On the event of Theorem~\ref{thm:mwu},
$\sum_t\lambda_t\ge T/3$, so $a\ge|\bar v|/3$. The number of attempts is geometric with mean $1/a\le3/|\bar v|$.

Now stop after $K$ attempts, and let $F$ be the event that all of them fail. The output is
the first successful attempt's output off $F$, and a draw from $q_1$ on $F$. By the previous paragraph, the output
given the complement of $F$ has law $\bar u$. So
\[
  \mu\;=\;(1-\Pr[F])\,\bar u+\Pr[F]\,q_1 .
\]
For every set $E\subseteq\X$ this gives $\mu(E)-\bar u(E)=\Pr[F]\bigl(q_1(E)-\bar u(E)\bigr)$, which lies in
$[-\Pr[F],\Pr[F]]$. So $\TV(\mu,\bar u)\le\Pr[F]$.

If $|\bar v|\ge\epsilon$, then $a\ge\epsilon/3$, and $\Pr[F]=(1-a)^K\le e^{-aK}\le\beta$. Hence $\TV(\mu,\bar u)\le\beta$. The
event of Theorem~\ref{thm:mwu} gives $|\bar v|\le1+\epsilon$ and $f_i(\bar v)\le\epsilon$, so Proposition~\ref{prop:normsafe} gives
$\TV(\bar u,q_i)\le\TV(p,q_i)+2\epsilon$. The triangle inequality for total variation then gives
$\TV(\mu,q_i)\le\TV(\mu,\bar u)+\TV(\bar u,q_i)\le\TV(p,q_i)+2\epsilon+\beta$.

If $|\bar v|<\epsilon$, then $\TV(\bar v,q_i)=\sum_x(q_i(x)-\bar v(x))_+\ge1-|\bar v|>1-\epsilon$. Together with $f_i(\bar v)\le\epsilon$ this gives
$\TV(p,q_i)>1-2\epsilon$. Every distribution is within $1$ of $q_i$, so $\mu$ satisfies the bound whatever its law.
\end{proof}

Given the stored rounds, the expected \OS{} and \OC{} costs per capped draw are at most $K$ times the corresponding
per-round bounds in Table~\ref{tab:cost}, plus one call to \OS{} for the fallback. The sampler makes no calls to \OP{}.

\begin{corollary}[Master theorem]\label{cor:master}
Suppose that Algorithm~\ref{alg:mwu} uses a proposal scheme that meets the requirements of Definition~\ref{def:scheme}. Then with
probability at least $1-\delta$, Algorithm~\ref{alg:mwu} returns a description of $\bar v$ with $|\bar v|\le1+\epsilon$ and $f_i(\bar v)\le\epsilon$ for all $i$, after
$T=O(\log(n/\delta)/\epsilon^2)$ rounds. For any $\beta\in(0,1)$ the capped sampler of Lemma~\ref{lem:output} then draws from a distribution $\mu$ with
$\TV(\mu,q_i)\le\TV(p,q_i)+2\epsilon+\beta$ for every $i$.
\end{corollary}

\begin{proof}
Take a round with $w_0<\frac12$, and condition on the past and on the scheme's setup, which fix the
iterate $v$. By the proposal property the race of that round is the race of Section~\ref{sec:race}, and by
requirement~(b) its subgradients at an accepted point are computable. Theorem~\ref{thm:race} gives $\Ex\hat g_i=\lambda\ell_i\ge\lambda f_i(v)$,
by Lemma~\ref{lem:wgain}(i), which is \Stwo{}. It also gives $\Ex\langle w,\hat g\rangle=\lambda\langle w,\ell\rangle$. Averaging over the setup
and using the bias condition gives $\Ex\langle w,\hat g\rangle\le\epsilon/12$ given the past, which is \Sone{}. Degenerate
rounds satisfy both conditions by the paragraph on them. Theorem~\ref{thm:mwu} and Lemma~\ref{lem:output} give the rest.
\end{proof}

The only failure event is that of Theorem~\ref{thm:mwu}. The binomial scheme runs its race with a sequential test,
which accepts slightly fewer proposals than the full test. Proposition~\ref{prop:binscheme}(b) shows that this race still
satisfies \Sone{} and \Stwo{} for a slightly thinner iterate $v'$. Its proof also gives $\Pr[\mathrm V,Y=x]=v'(x)/(1+|v'|)$ given the round's setup, as needed in Lemma~\ref{lem:output}.
So the conclusions of Corollary~\ref{cor:master} hold for it too.

The next three sections give the proposal schemes and count their oracle calls in one round.

\section{Uniform proposals}\label{sec:uniform}

This is the most direct proposal scheme. The reference vector is the true $\bar w$ and the kernel is the point rank $s^\ast$,
so the iterate is the exact best response, whatever its mass, and there is no bias at all. The challenge is
sampling a quantile. At each point we designate one index that carries the quantile value, and comparisons can identify it.
A uniform proposal is accepted with probability $|Q_{s^\ast}|/n$, and the test costs $n-1$ comparisons. So a round uses
at most $n$ proposals and $O(n^2)$ comparisons in expectation.

\paragraph{Designation.} Fix $x$. Order the indices by their value $q_i(x)$, breaking ties by index number. Let
$W_{<i}(x)$ be the $\bar w$-weight of the indices strictly before $i$. Call $i$ \emph{designated} at rank $s$ if
$W_{<i}(x)<s\le W_{<i}(x)+\bar w_i$. This mirrors \eqref{eq:sandwich}, with single indices in place of groups.

\begin{lemma}[Designation]\label{lem:desig}
At each $x$ and rank $s\in(0,1)$ exactly one index $j$ is designated, and $q_j(x)=Q_s(x)$. Testing whether a given
$i$ is designated costs $n-1$ comparisons.
\end{lemma}

\begin{proof}
The running total of weights in this order starts at $0$, ends at $1$ and crosses $s$ at exactly one index, the
first one at which it reaches $s$. Its value is the lower quantile. To test $i$, ask one comparison per $j\ne i$:
for $j<i$, $j$ comes first iff $q_j(x)>q_i(x)$ is false; for $j>i$, iff $q_i(x)>q_j(x)$ is true. Ties fall on the
correct side, since both strict comparisons fail.
\end{proof}

\begin{proposition}[The uniform scheme]\label{prop:uniform}
Let $\tilde w=\bar w$, $\Gamma=n$, and the point rank kernels $\pi_s=\delta_s$. At rank $s$ the proposal draws $I$ uniformly
from $[n]$ and $Y\sim q_I$, sets $r=s$, and accepts iff $I$ is designated at rank $s$ at $Y$. This proposal scheme meets the
requirements of Definition~\ref{def:scheme}, and its bias is zero: the kernel at $s^\ast$ has $\langle w,\ell\rangle=0$. At this rank the race
satisfies \Sone{} and \Stwo{} with no slack: $\Ex\langle w,\hat g\rangle=0$ and $\Ex\hat g_i\ge\lambda f_i(v)$.
\end{proposition}

\begin{proof}
\emph{(a) Proposal property.} By Lemma~\ref{lem:desig}, exactly one index $j$ is designated at $x$, and $q_j(x)=Q_s(x)$.
So $\Pr[Y=x,\text{accepted}]=\frac1nq_j(x)=Q_s(x)/n$. The rank is always $r=s$, so \eqref{eq:proposal} holds with $\pi_s=\delta_s$
and $\Gamma=n$.

\emph{(b) The accepted index.} An accepted $I$ is designated at $Y$, so $q_I(Y)=Q_r(Y)$, and $I$ lies in the group of
$Q_r(Y)$.

\emph{(c) Bias.} At the kernel at $s^\ast$ every rank equals $s^\ast$, so $s-s^\ast=0$. Since $\tilde w=\bar w$, also
$\eta_s=0$. So the bracket in \eqref{eq:wgain} vanishes, $\langle w,\ell\rangle=0$, and the bias condition~\eqref{eq:bias} holds.

Theorem~\ref{thm:race} gives \Stwo{} and $\Ex\langle w,\hat g\rangle=\lambda\langle w,\ell\rangle=0$.
\end{proof}

So Algorithm~\ref{alg:mwu} with uniform proposals runs MWU against the exact best response, with gains that are exactly
unbiased after the division by $1+|v|$.

\begin{remark}[Uniform, not weighted]\label{rem:unif}
It is tempting to propose $I\sim \bar w$. Then the acceptance probability must be divided by $\bar w_I$, up to a
constant, to keep the proposal property. Without that correction the accepted points can be biased, because the
weights already act through designation and would be counted twice. Designating a single index also accounts for ties.
Accepting every tied index would overcount.
\end{remark}

\begin{remark}[The factor $n$ for designated indices]\label{rem:tradeoff}
Let a proposal pick $i$ with probability $p_i$, draw $Y\sim q_i$, and accept with some probability $a_i(Y)\le1$ only when
$i$ is designated at $Y$. At a point $x$ where $i$ is designated and $q_i(x)>0$, the proposal property reads
$p_iq_i(x)a_i(x)=Q_{s^\ast}(x)/\Gamma=q_i(x)/\Gamma$. So $\Gamma\ge1/p_i$ for every $i$ designated somewhere with positive value.
If this holds for every $i$, then $\sum_ip_i=1$ forces $\Gamma\ge n$. For example, take uniform candidate weights, $s^\ast=1/n$,
and $q_i(i)=a$, $q_i(j)=(1-a)/(n-1)$ for $j\ne i$, with $n\ge2$ and $0<a<1/n$. Index $i$ is designated at $i$ and
$|Q_{s^\ast}|=na<1$, so the race needs at least $n/(1+na)>n/2$ proposals in expectation. The DKW scheme shrinks $n$ to $m$. Rank smoothing removes it.
\end{remark}

\paragraph{Cost per round.} A round makes at most $n$ proposals in expectation, by Lemma~\ref{lem:racelaw}. Each
costs one call to \OS{} and $n-1$ comparisons for designation. The subgradients cost $O(n)$ at the single stopping
point. So a round makes at most $n$ calls to \OS{}, at most $n(n-1)+O(n)$ comparisons, both in expectation, and at most
one call to \OP{}.

\section{Proposals from DKW-sparsified weights}\label{sec:dkw}

The uniform scheme pays a factor $n$ twice. A round uses at most $n$ proposals in expectation, and each designation
test costs $n-1$ comparisons. Both factors come from using all $n$ indices.
This section replaces $\bar w$ by the empirical distribution $\hat w$ of $m$ draws from it. That helps when $n$ is much
larger than the number of draws.

\paragraph{Intuition.} At each point, a quantile is a statement about the cumulative distribution of the values
$q_i(x)$. The empirical cumulative distribution is close to the true one, by the Dvoretzky--Kiefer--Wolfowitz
(DKW) inequality, with no dependence on $n$. The price is a bias. Under $\hat w$ the subgradients assign each point
a true rank that differs from the target rank by at most the DKW distance at that point. The target rank may be near
$1$, where a quantile can be much larger than the average $q_i(x)$. So we cannot bound the bias by comparing
quantiles at nearby ranks. We bound it using only the division by $1+|v|$, after which the iterate has mass at
most $1$. This bound gives a factor $\log(n/\epsilon)$ in $m$.

\begin{definition}[The DKW scheme]\label{def:dkwscheme}
Draw $m$ indices independently from $\bar w$. Let $\hat w_i$ be the fraction of draws equal to $i$, and let $S$ be the set
of indices drawn. Use $\tilde w=\hat w$, the point rank kernels $\pi_s=\delta_s$, and $\Gamma=|S|\le m$. At rank $s$ the proposal
draws $I$ uniformly from $S$ and $Y\sim q_I$, sets $r=s$, and accepts iff $I$ is designated at rank $s$ at $Y$ with respect
to $\hat w$.
\end{definition}

Lemma~\ref{lem:desig} applies verbatim with $\hat w$ and $S$ in place of $\bar w$ and $[n]$, so each designation test costs
$|S|-1$ comparisons. The subgradients $d_i(\cdot,s)$ are defined for every
$i\in[n]$, including $i\notin S$, by whether $q_i(x)$ lies below, at or above the quantile.

At a point $x$, let $\lambda_x$ be the largest absolute difference between the $\bar w$-weight and the $\hat w$-weight of a set of the form
$\{i:q_i(x)\le t\}$ or $\{i:q_i(x)<t\}$. This is the Kolmogorov--Smirnov distance at $x$.

\begin{lemma}[DKW at a point]\label{lem:dkw}
For every $x$ and $a>0$, $\Pr[\lambda_x>a]\le2e^{-2ma^2}$. Hence $\Ex\lambda_x<2/\sqrt m$.
\end{lemma}

\begin{proof}
The map $i\mapsto q_i(x)$ pushes $\bar w$ forward to a distribution on $\R$, and it pushes the $m$ draws forward to an
i.i.d.\ sample from it. So $\lambda_x$ is the Kolmogorov--Smirnov statistic of a one-dimensional sample, and the
DKW inequality with Massart's constant applies. Integrate the tail for the expectation.
\end{proof}

\begin{proposition}[DKW bias]\label{prop:dkwbias}
Suppose $w_0<\frac12$ and let the response be the point rank $s^\ast$ under $\hat w$, with iterate $\tilde u=Q^{\hat w}_{s^\ast}$ and $\lambda=1/(1+|\tilde u|)$. Then
for every $\tau\in(0,1]$,
\[
  \Ex\lambda\langle w,\ell\rangle\;\le\;\tau+2n\,e^{-2m\tau^2}+\frac2{\sqrt m},
\]
where the expectation is over the draws.
\end{proposition}

\begin{proof}
\emph{The rank perturbation.} Put $\tilde s(x)=\frac12\bigl(1+\sum_i\bar w_ig_i(x,s^\ast)\bigr)$, with $g_i$ taken under $\hat w$. This is the
true rank that the subgradients assign to the value $\tilde u(x)$. Lemma~\ref{lem:facts}(b) for $\hat w$ gives
$\sum_i\hat w_ig_i=2s^\ast-1$, so $\tilde s(x)-s^\ast=\frac12\eta_{s^\ast}(x)$. The kernel is the point rank at $s^\ast$, so \eqref{eq:wgain}
reads
\[
  \langle w,\ell\rangle\;=\;D\sum_x\bigl(\tilde s(x)-s^\ast\bigr)\bigl(\tilde u(x)-p(x)\bigr).
\]
We show that $|\tilde s(x)-s^\ast|\le\lambda_x$. Let $A$ be the indices below $\tilde u(x)$ and $T$ those tied with it, and
let $\theta=\frac12(1+\gamma_{s^\ast}(x))\in[0,1]$ for the tie value. Summing $g_i$ against a probability vector $\omega$ gives
$2(\omega(A)+\theta\omega(T))-1$. With $\omega=\bar w$ this is $2\tilde s(x)-1$, and with $\omega=\hat w$ it is $2s^\ast-1$. So
\[
  \tilde s(x)-s^\ast\;=\;(1-\theta)\bigl(\bar w(A)-\hat w(A)\bigr)+\theta\bigl(\bar w(A\cup T)-\hat w(A\cup T)\bigr),
\]
a convex combination of two differences of prefix sets. Each is at most $\lambda_x$ in absolute value.

\emph{The bias.} With $D\le1$ the identity gives $\langle w,\ell\rangle\le\sum_x\lambda_x(\tilde u(x)+p(x))$. Multiply by $\lambda$. The
measure $\lambda\tilde u$ has mass at most $1$, and $\tilde u(x)$ is one of the values $q_i(x)$. So
\[
  \lambda\sum_x\lambda_x\tilde u(x)\;\le\;\tau+\sum_{x:\,\lambda_x>\tau}\ \sum_{i=1}^nq_i(x),
\]
because $\lambda_x\le1$. The last sum does not depend on the rank. By Lemma~\ref{lem:dkw} its expectation is at most
$2ne^{-2m\tau^2}$, since $\sum_x\sum_iq_i(x)=n$. Finally $\lambda\sum_x\lambda_xp(x)\le\sum_x\lambda_xp(x)$, whose expectation is below
$2/\sqrt m$, because $p$ does not depend on the draws.
\end{proof}

\begin{corollary}[Sample size]\label{cor:dkwm}
With $\tau=\epsilon/36$ and
\[
  m\;=\;\Bigl\lceil\frac{1296\,\ln(72n/\epsilon)}{\epsilon^2}\Bigr\rceil,
\]
the DKW scheme meets the requirements of Definition~\ref{def:scheme}. So the race satisfies $\Ex\langle w,\hat g\rangle\le\epsilon/12$
and $\Ex[\hat g_i\mid\text{past},\hat w]\ge\lambda f_i(\tilde u)$ in every round with $w_0<\frac12$.
\end{corollary}

\begin{proof}
\emph{(a) Proposal property.} Given the draws, Lemma~\ref{lem:desig} with $\hat w$ and $S$ in place of $\bar w$ and $[n]$ shows that
exactly one index $j\in S$ is designated at $x$, with $q_j(x)=Q_s(x)$. So $\Pr[Y=x,\text{accepted}]=Q_s(x)/|S|$ and $r=s$, which
is \eqref{eq:proposal} with $\pi_s=\delta_s$ and $\Gamma=|S|$.

\emph{(b) The accepted index.} An accepted $I$ is designated, so $q_I(Y)=Q_r(Y)$ under $\hat w$.

\emph{(c) Bias.} Proposition~\ref{prop:dkwbias} bounds the left side of \eqref{eq:bias}, and each of its three terms is at
most $\epsilon/36$. Indeed $2m\tau^2\ge\ln(72n/\epsilon)$ gives $2ne^{-2m\tau^2}\le\epsilon/36$, and
$\ln(72n/\epsilon)\ge4$ gives $m\ge5184/\epsilon^2$, so $2/\sqrt m\le\epsilon/36$.

The race claims are Theorem~\ref{thm:race} and Lemma~\ref{lem:wgain}, since the subgradients are those of $\tilde u$ under
$\hat w$.
\end{proof}

The factor $\ln(72n/\epsilon)$ comes from the crude bound $\tilde u(x)\le\sum_iq_i(x)$ on the rare points with large
$\lambda_x$. The $\hat w$-weight strictly below $\tilde u(x)$ is less than $s^\ast$, so the $\bar w$-weight there is less than $s^\ast+\lambda_x$.
If $s^\ast\le\frac34$ and $\lambda_x\le\frac18$, the true quantile at rank $s^\ast+\lambda_x$ is therefore at least $\tilde u(x)$. By Lemma~\ref{lem:facts}(c) it is at most $8\sum_i\bar w_iq_i(x)$. These points contribute at most $16/\sqrt m$
in expectation to the iterate term. The remaining points contribute at most $2ne^{-m/32}$ by the same crude bound,
and the $p$ term is still at most $2/\sqrt m$. So $m=O(\epsilon^{-2}+\log n)$ suffices at these ranks.

\paragraph{Cost per round.} The $m$ draws use no oracle. A round makes at most $m$ proposals in expectation, each
with one call to \OS{} and $|S|-1\le m$ comparisons for designation. The subgradients of all $n$ experts at the
stopping point cost $O(n+m)$, and $m=O(\epsilon^{-2}\log\frac n\epsilon)$.

\section{Rank-smoothed proposals}\label{sec:smooth}

Both schemes so far use point ranks, with rejection constant equal to the number of indices they use.
Remark~\ref{rem:tradeoff} shows that this can be necessary for a designated-index proposal at a point rank. So we spread the rank over a kernel of
width $\tilde\epsilon=\Theta(\epsilon)$. We keep $\tilde w=\bar w$ throughout this section.

\paragraph{Intuition.} Draw $I\sim \bar w$ and $Y\sim q_I$. The proposal lands at some rank among the values $q_j(Y)$, and
we can make that rank uniform within ties. The key fact is that this rank has a clean law: the pair $(Y,r)$ has
density $Q_s(x)$ (Lemma~\ref{lem:ranklaw}). So if we accept with probability $\kappa(r)$, the accepted pairs have
density proportional to $\kappa(s)Q_s(x)$. This is the proposal property for the kernel with density proportional to $\kappa$. A kernel
of width $\tilde\epsilon$ in either scheme below accepts with probability of order $\tilde\epsilon$ times the mass of its iterate, so the rejection constant is
$O(1/\tilde\epsilon)$. The price is the smoothing bias of Lemma~\ref{lem:response}, which is
$O(\tilde\epsilon)$.

The two kernels differ in how they test the rank. The window kernel computes $r$ with a full scan of $2(n-1)$
comparisons. The binomial kernel never computes $r$. It flips coins with bias exactly $r$, at most two comparisons each,
and accepts if the number of heads lands in a window. Proposals with ranks far from the kernel's rank are usually rejected
after few coins. Near rank $1$ more proposals can have ranks close to the kernel's rank, but there the coins have small
variance, so fewer coins suffice.

\begin{lemma}[Rank law]\label{lem:ranklaw}
Draw $I\sim \bar w$, $Y\sim q_I$ and $U$ uniform on $(0,1)$. Let $c'=q_I(Y)$, and let $A$ and $W$ be the $\bar w$-weights of the
indices $j$ with $q_j(Y)<c'$ and with $q_j(Y)=c'$. Put $r=A+UW$. Then for every $x$ and Borel $E\subseteq(0,1)$,
$\Pr[Y=x,\ r\in E]=\int_EQ_s(x)\,ds$. Moreover, almost surely $Q_r(Y)=c'$, and $g_i(Y,r)$ equals $+1$, $2U-1$ or $-1$
according as $q_i(Y)$ is below, equal to or above $c'$.
\end{lemma}

\begin{proof}
Fix $x$. The indices with a common value $c''$ at $x$ form a group of weight $W'$, which occupies a rank interval of
length $W'$ on which $Q_s(x)=c''$. The event that $Y=x$ and $q_I(x)=c''$ has probability
$\sum_{i\text{ in group}}\bar w_iq_i(x)=W'c''$. Spreading this mass uniformly over the group's interval gives joint density
$c''=Q_s(x)$ there. On that interval $\gamma_r=(2r-2A-W)/W=2U-1$.
\end{proof}

So the rank of a proposal has density $|Q_s|$ at $s$, which integrates to $1$ and satisfies $|Q_s|\le1/(1-s)$,
by Lemma~\ref{lem:facts}. If we accept with probability $\kappa(r)$ for a known $\kappa:(0,1)\to[0,1]$ with
$Z=\int\kappa>0$, the proposal property holds with $\pi(dt)=\kappa(t)\,dt/Z$ and $\Gamma=1/Z$.

\paragraph{The smoothing bias.} Smoothing the rank creates slack in \Sone{}. Dividing by $1+|v|$ keeps that slack
small whatever the mass of the response. We may therefore center the kernel at the best-response rank
itself. We first turn the weighted gain \eqref{eq:wgain} into a form that shows this, and then bound it.

With $\tilde w=\bar w$ the term $\eta_s$ vanishes, and the density
$\psi_s=D(s-s^\ast)$ is the same at every point. For a constant $a$, $\langle a,Q_s-p\rangle=a(|Q_s|-|p|)=a(|Q_s|-1)$. So
\eqref{eq:wgain} becomes
\begin{equation}\label{eq:wgainref}
  \langle w,\ell\rangle\;=\;D\int(s-s^\ast)\bigl(|Q_s|-1\bigr)\,\pi(ds).
\end{equation}
We bound this gain when $\pi$ is concentrated near $s^\ast$. A family of kernels has \emph{radius} $\varrho$ and \emph{tail} $\vartheta$ if
each $\pi_s$ has density at most $\vartheta$ outside $[s-\varrho,s+\varrho]$. Both kernel families below have a kernel at every
rank in $(0,1)$, and $s^\ast\in(0,1)$ when $w_0<\frac12$. So we center the kernel at $s^\ast$ itself, even near rank $0$ or $1$.

\begin{lemma}[The scaled weighted gain]\label{lem:response}
Suppose $w_0<\frac12$ and take $\tilde w=\bar w$ and a family of radius $\varrho$ and tail $\vartheta$. Let the response be the
kernel at $s^\ast$, with iterate $v$ and $\lambda=1/(1+|v|)$. Then
\[
  \lambda\langle w,\ell\rangle\;\le\;\varrho+2\vartheta,
\]
whatever the mass $|v|$. In particular the bias condition~\eqref{eq:bias} holds if $\varrho+2\vartheta\le\epsilon/12$.
\end{lemma}

\begin{proof}
Use \eqref{eq:wgainref}, and bound $\bigl||Q_t|-1\bigr|$ by $|Q_t|+1$. On the main support of the kernel, $|t-s^\ast|\le\varrho$, so that
part of the integral is at most $\varrho\int(|Q_t|+1)\,\pi(dt)=\varrho(|v|+1)$. Outside it the density is at most $\vartheta$ and
$|t-s^\ast|\le1$, and $\int_0^1(|Q_t|+1)\,dt=2$ by Lemma~\ref{lem:facts}(d). So $\langle w,\ell\rangle\le D[\varrho(|v|+1)+2\vartheta]$. Multiply
by $\lambda$, and use $D\le1$, $\lambda(|v|+1)=1$ and $\lambda\le1$.
\end{proof}

\subsection{The window kernel}\label{sec:window}

The window kernel accepts exactly when the rank falls in a window of width $\tilde\epsilon$ around the kernel's rank.
Ranks lie in $(0,1)$, so the test cuts the window off at $0$ and $1$ by itself. Testing the rank needs
$A$ and $W$, which takes one full scan. The scan also yields every subgradient at the proposal, so accepted points
cost nothing extra.

\begin{proposition}[The window scheme]\label{prop:window}
Let $\tilde\epsilon=\epsilon/6$ and, for a rank $s\in(0,1)$, let $\kappa_s$ be the indicator of
$[s-\frac{\tilde\epsilon}2,s+\frac{\tilde\epsilon}2]\cap(0,1)$. Then $Z_s=\min\{1,s+\frac{\tilde\epsilon}2\}-\max\{0,s-\frac{\tilde\epsilon}2\}\ge\frac{\tilde\epsilon}2$, so
$\Gamma=1/Z_s\le2/\tilde\epsilon=12/\epsilon$. The proposal draws $I\sim\bar w$, $Y\sim q_I$ and $U$, computes $r$ as in
Lemma~\ref{lem:ranklaw}, and accepts iff $r\in[s-\frac{\tilde\epsilon}2,s+\frac{\tilde\epsilon}2]$. This proposal scheme meets the
requirements of Definition~\ref{def:scheme}, with $\lambda\langle w,\ell\rangle\le\frac{\tilde\epsilon}2=\epsilon/12$ for the kernel at $s^\ast$. So the
race satisfies \Stwo{} with no slack, and \Sone{}.
\end{proposition}

\begin{proof}
\emph{(a) Proposal property.} At least half of the window lies in $(0,1)$, since $s\in(0,1)$, so $Z_s\ge\frac{\tilde\epsilon}2$. The
test accepts with probability $\kappa_s(r)$, so Lemma~\ref{lem:ranklaw} gives \eqref{eq:proposal} with $\pi_s(dt)=\kappa_s(t)\,dt/Z_s$ and
$\Gamma=1/Z_s$. We know $Z_s$, so $\Gamma$ needs no oracle, and the $\mathrm P$ branch draws its rank uniformly from the cut-off
window.

\emph{(b) The accepted index.} By Lemma~\ref{lem:ranklaw}, $Q_r(Y)=q_I(Y)$ almost surely, so $I$ lies in the group of
$Q_r(Y)$.

\emph{(c) Bias.} The family has radius $\frac{\tilde\epsilon}2$ and tail $0$. Lemma~\ref{lem:response} gives
$\lambda\langle w,\ell\rangle\le\frac{\tilde\epsilon}2=\epsilon/12$.

Theorem~\ref{thm:race} and Lemma~\ref{lem:wgain} give the claims about the race.
\end{proof}

\paragraph{Cost per round.} To test a proposal $(I,Y)$, ask both comparisons between $I$ and every other $j$ at $Y$.
This gives $A$ and $W$, hence $r$, with $2(n-1)$ calls, and the same answers give all subgradients on acceptance.
With $\Gamma\le12/\epsilon$, a round makes at most $12/\epsilon$ calls to \OS{} and $24(n-1)/\epsilon+O(n)$ comparisons, in
expectation.

\subsection{The binomial kernel with early rejection}\label{sec:binomial}

The binomial kernel removes the full scan. A coin with bias exactly $r$ costs at most two comparisons, as we show below.
Flip $N$ such coins and accept if the fraction of heads lies within $\tilde\epsilon/4$ of the kernel's rank $s$. The
acceptance probability, as a function of $r$, is a smooth bump of width about $\tilde\epsilon$. Its normalizer is
known exactly, and its tails are exponentially small. The number of coins comes from the variance. A coin with bias
near $s$ has variance at most $s(1-s)+|r-s|$. Up to logarithmic factors, our count is
$s(1-s)/\tilde\epsilon^2+1/\tilde\epsilon$. A proposal at distance $d\ge2\tilde\epsilon$ from $s$ is usually
rejected after $O(\Lambda(s(1-s)+d)/d^2)$ coins, where $\Lambda=O(\log(1/\epsilon))$. The two
choices work together. Far from rank $1$, few proposals land near $s$, and early rejection makes the others cheap.
Near rank $1$, up to a fraction of about $\tilde\epsilon/(1-s)$ of the proposals lands near $s$, but each of them
costs $O(\Lambda((1-s)/\tilde\epsilon^2+1/\tilde\epsilon))$ coins. Either way a proposal costs
$O(\log(1/\epsilon)/\epsilon)$ coins on average. The sequential test can occasionally disagree with the full test,
and we treat the disagreement as one more source of bias.

\paragraph{The kernel.} Let $\tilde\epsilon=\epsilon/16$, $C_0=2^{16}$ and $\Lambda=2\ln(C_0/\tilde\epsilon^4)$. For a rank
$t\in(0,1)$ write $\sigma_t^2=t(1-t)$, the variance of a coin with bias $t$. For a rank $s\in(0,1)$ put
\[
  N_s=\Bigl\lceil C_0\Lambda\Bigl(\frac{\sigma_s^2}{\tilde\epsilon^2}+\frac1{\tilde\epsilon}\Bigr)\Bigr\rceil,\quad
  N_-=\max\bigl\{0,\bigl\lceil N_s(s-\tfrac{\tilde\epsilon}4)\bigr\rceil\bigr\},\quad
  N_+=\min\bigl\{N_s,\bigl\lfloor N_s(s+\tfrac{\tilde\epsilon}4)\bigr\rfloor\bigr\},
\]
and let $\kappa_s(t)$ be the probability that a binomial variable with $N_s$ trials and success probability $t$ lies
in $[N_-,N_+]$. The band is cut off at $0$ and $N_s$ only within $\frac{\tilde\epsilon}4$ of rank $0$ or $1$. At $s=\frac12$ the count $N_s$ is of order $\Lambda/\tilde\epsilon^2$. Within $O(\tilde\epsilon)$ of rank $0$
or $1$ it is only of order $\Lambda/\tilde\epsilon$.

\begin{lemma}[Properties of the binomial kernel]\label{lem:binom}
For every rank $s\in(0,1)$:
\begin{enumerate}
\item[(a)] $Z_s=(N_+-N_-+1)/(N_s+1)$ exactly, and $\tilde\epsilon/8\le Z_s\le\tilde\epsilon$, so $\Gamma\le8/\tilde\epsilon$. A draw from
$\pi_s$, with density $\kappa_s/Z_s$, needs no oracle.
\item[(b)] If $|t-s|\ge\tilde\epsilon$, then $\kappa_s(t)\le e^{-\Lambda}$. So the family has radius $\tilde\epsilon$ and tail
at most $\tilde\epsilon/8$.
\item[(c)] $\Lambda\ge1$, $e^{-\Lambda}=\tilde\epsilon^8/C_0^2\le\tilde\epsilon^2/64$ and $\Lambda e^{-\Lambda}\le\tilde\epsilon^4/C_0$. Also
$N_s\le C_0\Lambda/(3\tilde\epsilon^2)$ and $N_s\le C_0\Lambda((1-s)/\tilde\epsilon^2+1/\tilde\epsilon)+1$.
\end{enumerate}
\end{lemma}

\begin{proof}
\begin{enumerate}
\item[(a)] Each binomial probability integrates to $1/(N_s+1)$ over $t\in(0,1)$, by the beta integral. The allowed counts are
the integers in $[N_s(s-\frac{\tilde\epsilon}4),N_s(s+\frac{\tilde\epsilon}4)]\cap[0,N_s]$. Since $s\in(0,1)$, this interval has length at
least $N_s\tilde\epsilon/4\ge1$, because $N_s\ge4/\tilde\epsilon$. If it is cut off, an endpoint is the integer $0$ or $N_s$.
Otherwise its length is $N_s\tilde\epsilon/2$. Either way it contains at least $N_s\tilde\epsilon/4$ and at most $N_s\tilde\epsilon/2+1$ integers, and $N_s+1\le2N_s$. Also $\pi_s$ is the uniform mixture,
over allowed counts $k$, of the laws $\mathrm{Beta}(k+1,N_s-k+1)$. To draw from it, pick $k$ uniformly from the band
and take order statistic $k+1$ of $N_s+1$ independent uniforms, in $O(N_s)$ operations.
\item[(b)] Let $d=|t-s|$. Acceptance needs the fraction of heads to deviate from $t$ by at least $d-\tilde\epsilon/4\ge3d/4$,
in one direction. The coin variance is $\sigma_t^2\le\sigma_s^2+d$, because $|\sigma_t^2-\sigma_s^2|=|t-s|\,|1-t-s|$. Bernstein's
inequality with deviation $3d/4$ bounds the probability by $\exp(-9N_sd^2/(40(\sigma_s^2+d)))$. Since $d\ge\tilde\epsilon$,
the choice of $N_s$ gives $N_sd^2\ge C_0\Lambda(\sigma_s^2+d)$, so the exponent is at least $9C_0\Lambda/40\ge\Lambda$. Then (a)
and (c) give the tail bound.
\item[(c)] Put $\Xi=C_0/\tilde\epsilon^4$, so $\Lambda=2\ln\Xi$ and $e^{-\Lambda}=\Xi^{-2}$. Then $\Lambda e^{-\Lambda}=2\ln\Xi/\Xi^2\le1/\Xi$. Also $\tilde\epsilon^2\Lambda$ is increasing up to $\tilde\epsilon=1/16$, where it equals $(\ln2)/4<1$. With $\sigma_s^2\le\frac14$ and $\tilde\epsilon\le\frac1{16}$,
$N_s\le\frac{5C_0\Lambda}{16\tilde\epsilon^2}+1\le\frac{C_0\Lambda}{3\tilde\epsilon^2}$. The last bound uses $\sigma_s^2\le1-s$.
\end{enumerate}
\end{proof}

\paragraph{Coins.} A proposal is $(I,Y,U)$, with $c'=q_I(Y)$, $A$, $W$ and $r=A+UW$ as in Lemma~\ref{lem:ranklaw}. To
flip one coin, draw $J\sim \bar w$ and $V$ uniform, both internally. The coin is heads if $q_J(Y)<c'$, or if $q_J(Y)=c'$
and $V<U$. We never evaluate $c'$. The call $q_I(Y)>q_J(Y)$ settles the first case. If it returns false, the call
$q_J(Y)>q_I(Y)$ separates ``above'' (tails) from a tie (heads iff $V<U$). Given the proposal, the coin is heads with
probability $A+UW=r$. So each coin is an independent Bernoulli$(r)$ flip and costs at most two comparisons.
Accepting iff the number of heads among $N_s$ coins lies in $[N_-,N_+]$ accepts with probability exactly $\kappa_s(r)$.
On acceptance, Lemma~\ref{lem:ranklaw} gives all subgradients from one scan against $I$.

\paragraph{Early rejection.} We use thresholds that adapt to the variance of the
coins. Fix a proposal with rank $r$, put $d=|r-s|$, and let $\hat r_\tau$ be the fraction of heads among the first
$\tau$ coins. Put
\[
  b_\tau(r)=\sqrt{\frac{2\sigma_r^2\Lambda}\tau}+\frac\Lambda\tau,\qquad
  B_\tau=\sqrt{\frac{2(\sigma_s^2+\tilde\epsilon)\Lambda}\tau}+\frac\Lambda\tau.
\]
The radius $b_\tau(r)$ is a Bernstein confidence radius for $\hat r_\tau$, and it depends on the unknown $r$. The
threshold $B_\tau$ is known. It bounds $b_\tau(r)$ whenever $d\le\tilde\epsilon$, because then $\sigma_r^2\le\sigma_s^2+\tilde\epsilon$.

\begin{definition}[Sequential test]\label{def:seq}
Flip coins one at a time. At each checkpoint $\tau$ that is a power of two below $N_s$, reject and stop if
$|\hat r_\tau-s|\ge\tilde\epsilon+B_\tau$. If no checkpoint rejects, flip coins up to $N_s$ and apply the full test.
\end{definition}

For one proposal, let $\Nflip$ be the number of coins that the sequential test flips before it accepts or rejects. It is
random, it depends on the rank $r$ of the proposal, and $\Nflip\le N_s$ always. Each coin costs at most two comparisons,
so $\Nflip$ controls the comparison cost of one acceptance test.

The sequential test reads a prefix of the full test's coins, so we compare the two on the same coins. Let $\mathcal E$
be the event that $|\hat r_\tau-r|<b_\tau(r)$ at every checkpoint. The radius satisfies
$\tau b_\tau(r)^2\ge2\Lambda(\sigma_r^2+b_\tau(r)/3)$. So Bernstein's inequality and a union bound over the at most
$\lceil\log_2N_s\rceil$ checkpoints give $\Pr[\text{not }\mathcal E\mid r]\le\zeta_0\coloneqq2\lceil\log_2N_s\rceil e^{-\Lambda}$.

\begin{lemma}[Disagreement and stopping time]\label{lem:seq}
Given $r$:
\begin{enumerate}
\item[(i)] the sequential test accepts only if the full test accepts, and the two disagree with probability at
most $\zeta\coloneqq\zeta_0+e^{-\Lambda}$;
\item[(ii)] if $d\ge2\tilde\epsilon$, then on $\mathcal E$ we have $\Nflip\le\min\{N_s,128\Lambda(\sigma_s^2+d)/d^2\}$.
\end{enumerate}
\end{lemma}

\begin{proof}
\begin{enumerate}
\item[(i)] The sequential test accepts only by running the full test, so a disagreement is an early rejection where the
full test would accept. Suppose that $\mathcal E$ holds and $d\le\tilde\epsilon$. Then $b_\tau(r)\le B_\tau$, so
$|\hat r_\tau-s|<d+B_\tau\le\tilde\epsilon+B_\tau$ at every checkpoint, and no checkpoint rejects. So on $\mathcal E$ an early
rejection forces $d>\tilde\epsilon$. Then the full test accepts with probability $\kappa_s(r)\le e^{-\Lambda}$, by
Lemma~\ref{lem:binom}(b). Add the probability of leaving $\mathcal E$.
\item[(ii)] Since $\sigma_r^2\le\sigma_s^2+d$ and $\tilde\epsilon<d$, both $b_\tau(r)$ and $B_\tau$ are at most
$\sqrt{2(\sigma_s^2+d)\Lambda/\tau}+\Lambda/\tau$. Take a checkpoint $\tau\ge64\Lambda(\sigma_s^2+d)/d^2$. Then this bound is at
most $d/\sqrt{32}+d/64<d/5$. On $\mathcal E$, $|\hat r_\tau-s|\ge d-b_\tau(r)>\frac45d$. Since $d\ge2\tilde\epsilon$, also
$\frac45d\ge\tilde\epsilon+\frac15d>\tilde\epsilon+B_\tau$. So the test rejects at $\tau$. The first such checkpoint is below
$128\Lambda(\sigma_s^2+d)/d^2$. If there is none, then $N_s$ is below that bound too.
\end{enumerate}
\end{proof}

\begin{proposition}[Coins per proposal]\label{prop:coins}
At every kernel rank $s$, $\Ex\Nflip\le(10C_0+512)\Lambda/\tilde\epsilon+3\le11C_0\Lambda/\tilde\epsilon$. Always
$\Nflip\le N_s\le C_0\Lambda/(3\tilde\epsilon^2)$.
\end{proposition}

\begin{proof}
The argument uses only $N_s$, the thresholds of the sequential test and the rank density, not the band, so it holds
at every $s\in(0,1)$. Let $\gamma=1-s$, so that $\sigma_s^2\le\gamma$. If $\gamma<4\tilde\epsilon$, then $\Nflip\le N_s\le5C_0\Lambda/\tilde\epsilon+1$, by
Lemma~\ref{lem:binom}(c). So suppose $\gamma\ge4\tilde\epsilon$. We use $\Nflip\le N_s$ when $d<2\tilde\epsilon$ or off $\mathcal E$, and
Lemma~\ref{lem:seq}(ii) otherwise. The rank has density $|Q_r|\le1/(1-r)$, by Lemmas~\ref{lem:ranklaw}
and~\ref{lem:facts}(c). We split the expectation into five parts.
\begin{itemize}[topsep=2pt,itemsep=1pt]
\item \emph{Ranks near $s$.} $\Pr[d<2\tilde\epsilon]\le4\tilde\epsilon/(\gamma-2\tilde\epsilon)\le8\tilde\epsilon/\gamma$. By
Lemma~\ref{lem:binom}(c), $N_s\cdot8\tilde\epsilon/\gamma\le8C_0\Lambda(1/\tilde\epsilon+1/\gamma)+8\tilde\epsilon/\gamma\le10C_0\Lambda/\tilde\epsilon+2$.
\item \emph{Off $\mathcal E$.} $N_s\zeta_0\le2N_s^2e^{-\Lambda}\le\frac29\Lambda^2\tilde\epsilon^4\le1$, by Lemma~\ref{lem:binom}(c) and
$\Lambda\tilde\epsilon^2\le1$ (the left side increases up to $(\ln 2)/4$ at $\tilde\epsilon=1/16$).
\item \emph{Ranks below $s-2\tilde\epsilon$.} At $r=s-d$ the density is at most $1/(\gamma+d)$, and $\sigma_s^2+d\le\gamma+d$. So
these ranks contribute at most $128\Lambda\int_{2\tilde\epsilon}^\infty u^{-2}du=64\Lambda/\tilde\epsilon$.
\item \emph{Ranks from $s+2\tilde\epsilon$ to $s+\frac\gamma2$.} The density is at most $2/\gamma$, and
$\sigma_s^2+d\le\frac32\gamma$. So these ranks contribute at most $384\Lambda\int_{2\tilde\epsilon}^\infty u^{-2}du=192\Lambda/\tilde\epsilon$.
\item \emph{Ranks above $s+\frac\gamma2$.} Here $d>\frac\gamma2$ and $\sigma_s^2+d\le2\gamma$, so
$\Nflip\le1024\Lambda/\gamma\le256\Lambda/\tilde\epsilon$. These ranks have total probability at most $1$.
\end{itemize}
The sum is the first bound. The second uses $C_0\ge515$, and also covers the case $\gamma<4\tilde\epsilon$.
\end{proof}

So a proposal costs $O(\Lambda/\tilde\epsilon)$ coins in expectation at every kernel rank $s$. Far from rank $1$ the rank density is
bounded near $s$. Only an $O(\tilde\epsilon)$ fraction of the proposals has $d<2\tilde\epsilon$ there. On $\mathcal E$ the others stop after
$O(\Lambda(\sigma_s^2+d)/d^2)$ coins. Near rank $1$ the fraction with $d<2\tilde\epsilon$ can be as large as $O(\tilde\epsilon/(1-s))$,
but there $N_s$ is only $O(\Lambda((1-s)/\tilde\epsilon^2+1/\tilde\epsilon))$.

\begin{remark}[A fixed count is too expensive]\label{rem:fixedN}
Suppose every kernel used the count of rank $\frac12$, $N=\lceil C_0\Lambda(\frac1{4\tilde\epsilon^2}+\frac1{\tilde\epsilon})\rceil$,
with the Hoeffding thresholds $\tilde\epsilon+\sqrt{\Lambda/(2\tau)}$. Take $k=\lceil16/\tilde\epsilon\rceil$ distributions with disjoint supports
and uniform relative weights. Choose $w_0$ so that $s^\ast=1-1/k-\tilde\epsilon/2$. Every proposal has rank in $[1-1/k,1]$,
so $\tilde\epsilon/2\le r-s^\ast\le9\tilde\epsilon/16$. By Hoeffding, except with probability $2\lceil\log_2N\rceil e^{-\Lambda}$,
$|\hat r_\tau-r|<\sqrt{\Lambda/(2\tau)}$ at every checkpoint. Then the proposal flips all $N$ coins.
Yet the full test accepts with probability at most $e^{-N\tilde\epsilon^2/8}$, since $r$ lies at least $\tilde\epsilon/4$ above its band.
As $1/\Gamma=\Theta(\tilde\epsilon)$, the race takes $\Theta(1/\tilde\epsilon)$ proposals in expectation. So this choice of weights costs
$\Theta(N/\tilde\epsilon)$ coin flips per round in expectation. With the adaptive count, $N_s=O(\Lambda/\tilde\epsilon)$ at these ranks,
and the round costs $\tilde O(\epsilon^{-2})$ comparisons, including the $O(k)$ comparisons at the stopping point.
\end{remark}

\begin{proposition}[The binomial scheme]\label{prop:binscheme}
We can run the binomial race with either test.
\begin{enumerate}
\item[(a)] With the full test, the binomial kernel gives a proposal scheme that meets the requirements of
Definition~\ref{def:scheme}, with $\lambda\langle w,\ell\rangle\le\frac54\tilde\epsilon<\epsilon/12$ for the kernel at $s^\ast$.
\item[(b)] Run the race for the binomial kernel at $s=s^\ast$ with the sequential test. Let $\kappa'_s(r)$ be the probability that the
sequential test accepts a proposal of rank $r$, and let $v'=\Gamma\int Q_t\kappa'_s(t)\,dt$ and $\lambda'=1/(1+|v'|)$. Then
$\Ex\langle w,\hat g\rangle<1.28\tilde\epsilon<\epsilon/12$, and $\Ex\hat g_i\ge\lambda'f_i(v')-\Gamma\zeta$ with $\Gamma\zeta\le\tilde\epsilon/100$. So \Sone{} and
\Stwo{} hold for the iterate $v'$.
\end{enumerate}
\end{proposition}

\begin{proof}
\emph{(a) The full test.} The coins accept a proposal of rank $r$ with probability $\kappa_s(r)$. So Lemma~\ref{lem:ranklaw}
gives the proposal property with $\pi_s(dt)=\kappa_s(t)\,dt/Z_s$ and $\Gamma=1/Z_s\le8/\tilde\epsilon$, by Lemma~\ref{lem:binom}(a). On
acceptance $Q_r(Y)=q_I(Y)$, by Lemma~\ref{lem:ranklaw}, so $I$ lies in the group of $Q_r(Y)$ and its tie value is $2U-1$. The family has radius $\tilde\epsilon$
and tail at most $\tilde\epsilon/8$, by Lemma~\ref{lem:binom}(b). So Lemma~\ref{lem:response} gives
$\lambda\langle w,\ell\rangle\le\tilde\epsilon+\frac{\tilde\epsilon}4=\frac54\tilde\epsilon$, which is below $\epsilon/12$ since $\tilde\epsilon=\epsilon/16$.

\emph{(b) The sequential test.} Let $\pi(dt)=\Gamma\kappa_s(t)\,dt$, $\pi'(dt)=\Gamma\kappa'_s(t)\,dt$ and $\nu=\pi-\pi'$. By Lemma~\ref{lem:seq}(i),
$0\le d\nu/dt=\Gamma(\kappa_s(t)-\kappa'_s(t))\le\Gamma\zeta$, so its mass $m_\nu=\nu((0,1))$ is at most $\Gamma\zeta$. The accepted pairs satisfy \eqref{eq:proposal} with $\pi'$ in place of $\pi$, so
Lemma~\ref{lem:racelaw} holds with $v'=\int Q_t\,\pi'(dt)$ and $\lambda'$.

\emph{The iterate and \Stwo{}.} The measure $\pi'$ has mass $1-m_\nu$. So $v'$ is the mixture of the quantiles $Q_t$, with
weights $\pi'$, and of the zero function, with weight $m_\nu$. Its linear gains are $\ell'_0=|v'|-1$ and
$\ell'_i=\int\langle d_i(\cdot,t),Q_t-p\rangle\,\pi'(dt)+m_\nu$, since $-1$ is a subgradient of $f_i$ at $0$ and
$\langle-1,0-p\rangle=1$. By convexity, as in Lemma~\ref{lem:wgain}(i), $\ell'_i\ge f_i(v')$. In the race, $\Ex\hat g_0=\lambda'\ell'_0$, and
\[
  \Ex\hat g_i=\lambda'\Bigl(\int\langle d_i,Q_t\rangle\,\pi'(dt)-\int\langle d_i,p\rangle\,\pi(dt)\Bigr)
  =\lambda'\ell'_i-\lambda'\Bigl(m_\nu+\int\langle d_i(\cdot,t),p\rangle\,\nu(dt)\Bigr),
\]
because the $\mathrm P$ branch draws its rank from $\pi$. The last bracket lies in $[0,m_\nu]$, since $d_i\in[-1,0]$. So
$\Ex\hat g_i\ge\lambda'f_i(v')-\Gamma\zeta$.

\emph{\Sone{}.} Let $v$ and $\ell$ be the iterate and gains of the full kernel $\pi$. Then $|v|-|v'|=\int|Q_t|\,\nu(dt)\in[0,\Gamma\zeta]$,
because $\int_0^1|Q_t|\,dt=1$. Compared with $\langle w,\ell\rangle$, the weighted gain of the race loses $w_0(|v|-|v'|)\ge0$ and
gains $-\sum_{i=1}^nw_i\int\langle d_i,Q_t\rangle\,\nu(dt)\le D\Gamma\zeta$. So $\Ex\langle w,\hat g\rangle\le\lambda'(\langle w,\ell\rangle+\Gamma\zeta)$. The proof of
Lemma~\ref{lem:response}, with radius $\tilde\epsilon$ and tail at most $\tilde\epsilon/8$, gives
$\langle w,\ell\rangle\le\tilde\epsilon(|v|+1)+\frac{\tilde\epsilon}4$, and $\lambda'(|v|+1)\le1+\Gamma\zeta$. Hence
$\Ex\langle w,\hat g\rangle\le\tilde\epsilon(1+\Gamma\zeta)+\frac{\tilde\epsilon}4+\Gamma\zeta$.

\emph{The slack.} We have $\zeta\le3N_se^{-\Lambda}$, and Lemma~\ref{lem:binom}(a) and (c) give $\Gamma\le8/\tilde\epsilon$,
$N_s\le C_0\Lambda/(3\tilde\epsilon^2)$ and $e^{-\Lambda}=\tilde\epsilon^8/C_0^2$. So $\Gamma\zeta\le8\Lambda\tilde\epsilon^5/C_0\le\tilde\epsilon/100$, using
$\Lambda\tilde\epsilon^2\le1$. This gives $\Ex\langle w,\hat g\rangle<1.28\tilde\epsilon$.
\end{proof}

\paragraph{Cost per round.} Here $\Gamma\le8/\tilde\epsilon=128/\epsilon$ and $\Lambda=O(\log(1/\epsilon))$. A round makes at most $128/\epsilon$
calls to \OS{} in expectation. The coins of one proposal cost at most $2\Ex\Nflip\le22C_0\Lambda/\tilde\epsilon$ comparisons in
expectation, at every kernel rank $s$, by Proposition~\ref{prop:coins}, and Wald's identity multiplies this by the expected number
of proposals. Add $O(n)$ for the subgradients at the stopping point. So a round makes
$O(\epsilon^{-2}\log\frac1\epsilon+n)$ comparisons in expectation.

\begin{remark}[Without early rejection]\label{rem:noER}
If every proposal flips all $N_s$ coins, a round uses $\Theta(\epsilon^{-3}\log(1/\epsilon))$ coins in expectation when the kernel's rank stays
bounded away from $0$ and $1$. Indeed $|v_s|=O(1)$ by Lemma~\ref{lem:facts}(c) within radius $\tilde\epsilon$ and the tail bound outside,
using $\int_0^1|Q_t|\,dt=1$. Early rejection improves the worst-case bound by a factor of order $1/\tilde\epsilon$. Within $O(\tilde\epsilon)$ of rank
$1$ the count $N_s$ is already $O(\Lambda/\tilde\epsilon)$, so there the adaptive count alone suffices.
\end{remark}

\section{Oracle complexity}\label{sec:cost}

Every proposal scheme runs $T=\lceil900\ln(2(n+1)/\delta)/\epsilon^2\rceil=O(\log(n/\delta)/\epsilon^2)$ rounds of Algorithm~\ref{alg:mwu}, by
Corollary~\ref{cor:master} and Proposition~\ref{prop:binscheme}(b), and makes at most one call to \OP{} per round. Table~\ref{tab:cost} lists the costs, in
expectation.

\begin{table}[ht]
\centering
\small
\begin{tabular}{@{}lllll@{}}
\toprule
Scheme & \OS{} per round & \OC{} per round & \OS{} total & \OC{} total \\
\midrule
Uniform & $O(n)$ & $O(n^2)$ & $O(n\epsilon^{-2}\log\frac n\delta)$ & $O(n^2\epsilon^{-2}\log\frac n\delta)$ \\
DKW & $O(\epsilon^{-2}\log\frac n\epsilon)$ & $O(\epsilon^{-4}\log^2\frac n\epsilon+n)$ & $O(\epsilon^{-4}\log\frac n\epsilon\log\frac n\delta)$ & $O((\epsilon^{-6}\log^2\frac n\epsilon+n\epsilon^{-2})\log\frac n\delta)$ \\
Window & $O(\epsilon^{-1})$ & $O(n\epsilon^{-1})$ & $O(\epsilon^{-3}\log\frac n\delta)$ & $O(n\epsilon^{-3}\log\frac n\delta)$ \\
Binomial & $O(\epsilon^{-1})$ & $O(\epsilon^{-2}\log\frac1\epsilon+n)$ & $O(\epsilon^{-3}\log\frac n\delta)$ & $O((\epsilon^{-4}\log\frac1\epsilon+n\epsilon^{-2})\log\frac n\delta)$ \\
\bottomrule
\end{tabular}
\caption{Oracle calls per round and in total, in expectation. The
binomial costs hold at every kernel rank, including ranks near $1$. Every proposal scheme makes at most $T=O(\epsilon^{-2}\log\frac n\delta)$
calls to \OP{}.}
\label{tab:cost}
\end{table}

\paragraph{High probability.} The number of proposals is random, but it concentrates. Every trial of a race stops
with probability at least $1/(2\Gamma_{\max})$, whatever happened before, where $\Gamma_{\max}$ is a fixed upper bound on the rejection constants.
So the total number of trials over all rounds is stochastically dominated by the number of independent coins
of bias $1/(2\Gamma_{\max})$ needed for $T$ heads. A Chernoff bound shows that it exceeds $M=\lceil4\Gamma_{\max}T\rceil$ with probability at
most $e^{-T/8}\le\delta/2$. For the binomial scheme take the first $M$ trials, padding with zeros if the run ends sooner.
The number of binomial coins in a trial is bounded by $B=C_0\Lambda/(3\tilde\epsilon^2)$ and has mean at most $b=11C_0\Lambda/\tilde\epsilon$ given the past before the trial.
Its conditional variance is at most $Bb$. So Freedman's inequality bounds the sum by $2Mb+O(B\log(2/\delta))$ except with probability $\delta/2$.
Adding $O(nT)$ comparisons at the stopping points gives the binomial totals with probability at least $1-\delta$.
Use $\delta/2$ for each guarantee to get accuracy and cost together.

\paragraph{Reading the table.} A round's calls to \OS{} are its proposals. Its calls to \OC{} are the proposals times the
cost of one acceptance test, plus $O(n)$ for the subgradients at the stopping point. The schemes trade the number of
proposals against the cost of a test. The totals multiply the per-round costs by $T=O(\epsilon^{-2}\log\frac n\delta)$.
\begin{itemize}
\item The uniform scheme makes $O(n)$ proposals per round, and each test costs $O(n)$ comparisons. That gives $O(n^2)$
comparisons per round. It has no bias.
\item The DKW scheme replaces both factors of $n$ by the number of draws, $O(\epsilon^{-2}\log\frac n\epsilon)$. It beats the
uniform scheme when $n\gg\epsilon^{-2}\log\frac n\epsilon$.
\item The window scheme makes $O(1/\epsilon)$ proposals per round, but each test still costs $O(n)$ comparisons. Its merit
is simplicity.
\item The binomial scheme makes $O(1/\epsilon)$ proposals per round, and a test costs $O(\epsilon^{-1}\log\frac1\epsilon)$ comparisons
on average, at every kernel rank. Up to logarithmic factors, it is the cheapest scheme when $n\gg1/\epsilon$.
\end{itemize}
We know $n$ and $\epsilon$ in advance. We use the uniform scheme when $n\le1/\epsilon$. For larger $n$ we use whichever of the
window and binomial schemes has the smaller comparison bound in the table. Both make $O(1/\epsilon)$ proposals per round, and
the window scheme makes $O(n/\epsilon)$ comparisons, with $n/\epsilon\le n^2$ there. So the total number of comparisons in expectation is
$O(\min\{n^2\epsilon^{-2},\epsilon^{-4}\log\frac1\epsilon+n\epsilon^{-2}\}\log\frac n\delta)$.

\section{AI Use Disclosure}
All parts of this version were created via an interactive process in which AI
(Claude Opus 5.5; ChatGPT Sol/Astra) generated drafts, details of proofs, and improved authors’
proofs based on author-provided structure and instructions. The resulting text was also extensively
edited and proofread by the authors. The authors take full responsibility for the content of this
paper.

\end{document}